\RequirePackage{fix-cm}
\documentclass[smallextended]{svjour3}       
\smartqed  
\usepackage{graphicx}
\usepackage{amsmath,algorithm,algpseudocode,epstopdf,graphicx,caption,subcaption,setspace,colortbl}
\usepackage{multirow,amsmath,amssymb,url}

\algnewcommand{\Inputs}[1]{%
	\State \textbf{Inputs:}
	\Statex \hspace*{\algorithmicindent}\parbox[t]{.8\linewidth}{\raggedright #1}
}
\algnewcommand{\Outputs}[1]{%
	\State \textbf{Outputs:}
	\Statex \hspace*{\algorithmicindent}\parbox[t]{.8\linewidth}{\raggedright #1}
}

\DeclareMathOperator*{\argmax}{arg\,max}
\DeclareMathOperator*{\argmin}{arg\,min}

\begin{document}

\title{Classification with Boosting of Extreme Learning Machine Over Arbitrarily Partitioned Data}


\author{Ferhat \"{O}zg\"{u}r \c{C}atak}


\institute{Ferhat \"{O}zg\"{u}r \c{C}atak \at
              T\"{U}B\.{I}TAK B\.{I}LGEM, Cyber Security Institute \\
              Kocaeli, Turkey\\
              Tel.: +90-262-6481000\\
              Fax: +90-262-6481100\\
              \email{ozgur.catak@tubitak.gov.tr}
}

\date{04.21.2015 / Accepted: date}

\maketitle

\begin{abstract}
Machine learning based computational intelligence methods are widely used to analyze large scale data sets in this age of big data. Extracting useful predictive modeling from these types of data sets is a challenging problem due to their high complexity. Analyzing large amount of streaming data that can be leveraged to derive business value is another complex problem to solve. With high levels of data availability (\textit{i.e. Big Data}) automatic classification of them has become an important and complex task. Hence, we explore the power of applying MapReduce based Distributed AdaBoosting of Extreme Learning Machine (ELM) to build a predictive bag of classification models. Accordingly, (i) data set ensembles are created; (ii) ELM algorithm is used to build weak learners (classifier functions); and (iii) builds a strong learner from a set of weak learners. We applied this training model to the benchmark knowledge discovery and data mining data sets.
\end{abstract}

\section{Introduction}
\label{intro}
It is clear that there has been an unexpected increase in the quantity and variety of data generated worldwide by computers, mobile phones, and sensors. Just as computer technology evolved, the quantity and variety of data has also increased, becoming more focused on storing every type of data, the so-called Big Data. As the volume of data to build a predictive model increases, the complexity of that training increases too. As a result, building actionable predictive modeling of a large scale unstructured data set is a definitive Big Data problem. Predictive learning models try to discover patterns of training data and label new data instances to the correct output value. To efficiently handle unstructured large scale big data sets, it is critical to develop new machine learning methods that combine several boosting and classification algorithms.

Extreme Learning Machine (ELM) was proposed by \cite{Huang2006489} based on generalized Single-hidden Layer Feedforward Networks (SLFNs). Main characteristics of ELM are small training time compared to traditional gradient-based learning methods, high generalization property of predicting unseen examples with multi-class labels and parameter free with randomly generated hidden nodes. ELM algorithm is used in many different areas including document classification \cite{zhao2011xml}, bioinformatics \cite{Wang2008262} multimedia recognition \cite{Zong20112541,lan2013extreme}.

In recent years, much computational intelligence research has been devoted to building predictive modeling of distributed and parallel frameworks. In this research, the proposed learning model creates data chunks with varying size and bag of classifier functions using ELM algorithm trained with these arbitrary chosen sub data set with AdaBoosting method for large scale predictions. By creating data chunks from the training data set using the MapReduce paradigm, each subset of the training data set is used to find out the set of ELM ensembles as a single global classifier function. 

The main objective of this work is to train large scale data sets using ELM and AdaBoost. Another objective is to achieve the model's classification performance with same or close to the conventional ELM method. Conventional ELM training cannot be applied to large scale data sets on a single computer because of their complexity. Then experiments section is split into two subsections: "commonly used data sets" in Section \ref{sec:commonds} and "large scale data sets" in Section \ref{sec:commonlargeds}. Commonly used data sets are suitable for training on a single computer with the conventional ELM algorithm. We trained these data sets both conventional and proposed methods to show the classification performance changes of the proposed method. Classification performance results are shown in Section \ref{sec:conv_elm}.

The contributions of this paper are as follows:
\begin{itemize}
	\item A generative MapReduce technique based AdaBoosted ELM classification model is proposed for learning, and thus, faster classification model training is achieved.
	\item This research proposes a new learning method for AdaBoosted ELM that achieves parallelization both in large scale data sets and reduced computational time of learning algorithm.
	\item Training computations of working nodes are independent from each other thus minimizing the data communication. The other approaches, including Support Vector Machine training need data communication for the support vector exchange. \cite{4470008,sun2012study,cloudsvm}.
\end{itemize}
The rest of the paper is organized as follows: Section \ref{sec:related} briefly introduces some of the earlier works related to our problem. Section \ref{sec:preliminaries} describes algorithm ELM, AdaBoost and MapReduce technique.  Section \ref{sec:approach} and Section \ref{sec:experiments} evaluates the proposed learning model. Section \ref{sec:conclusion} concludes this paper.
\section{Related work}\label{sec:related}
In this section, we describe the general overview of literature review. Section \ref{sec:litoverview} describes the general distributed ELM methods. Section \ref{sec:mr-elm-methods} shows the MapReduce based ELM training methods. 
\subsection{Literature Review Overview}\label{sec:litoverview}
MapReduce based learning algorithms from distributed data chunks has been studied by many researchers. Many different MapReduce based learning solutions over arbitrary partitioned data have been proposed recently. Some popular MapReduce based solutions to train machine learning algorithms in the literature include the following. Panda et al. proposed a learning tree model which is based on series of distributed computations, and implements each one using the MapReduce model of distributed computation \cite{Panda:2009:PMP:1687553.1687569}. Zhang et al. develops some algorithms using MapReduce to perform parallel data joins on large scale data sets \cite{Zhang:2012:EPK:2247596.2247602}. Sun et al. use batch updating based hierarchical clustering to reduce computational time and data communication  \cite{5372757}. Their approach uses co-occurence based feature selection to remove noisy features and decrease the dimension of the feature vectors. He et al. proposed parallel density based clustering algorithm (DBSCAN). They developed a partitioning strategy for large scale non-indexed data with a 4-stages MapReduce paradigm \cite{6121313}. Zhao et al. proposed parallel k-means clustering based on MapReduce \cite{zhao2009}. Their approaches focus on implementing k-means with the read-only convergence heuristic in the MapReduce pattern.
\subsection{MapReduce Based ELM Training Methods}\label{sec:mr-elm-methods}
Section \ref{sec:elm-star} - Section \ref{sec:elm-mr} describe five different MapReduce training methods of ELM algorithm. 
\subsubsection{ELM $\star $}\label{sec:elm-star}
Xin et al. proposed MapReduce based ELM training method called as ELM $^\ast $ \cite{elmstar}. Main idea behind this method is to calculate matrix multiplication of ELM to find weight vector. They show that Moore-Penrose generalized inverse operator is the most expensive computation part of the algorithm. As we know, matrix multiplication can be divide into smaller part. Using this property, they proposed an efficient implementation of training phase to manage massive data sets. The final output of this method is a single classifier function. In this paper, they proposed two different versions of ELM $^\ast $, naive and improved. In naive-ELM $^\ast $, the algorithm has two classes, Class Mapper and Class Reducer. Both classes contain only one method. In improved ELM $^\ast $, they decompose the calculation of matrix multiplication using MapReduce framework. Moreover, the proposed algorithm decreases the computation and communication cost. In the experimental platform, they used their synthetic data sets to evaluate the performance of the proposed algorithms with MapReduce framework. 
\subsubsection{OS-ELM based Classification in Hierarchical P2P Network}\label{sec:oselm}
Sun et al. proposed OS-ELM \cite{4012031} based distributed ensemble classification in P2P networks \cite{Sun20112438}. They apply the incremental learning principle of OS-ELM to hierarchical P2P network. They proposed two different versions of the ensemble classifier in hierarchical P2P, \textit{one-by-one} ensemble classification and \textit{parallel} ensemble classification. In \textit{one-by-one} learning method, each peer, one by one, calculates the classifier with all the data. Therefore, this approach has a large network delay. In the \textit{parallel} ensemble learning, all the classifiers are learnt from all the data in parallel manner. Conversely to ELM $^\ast $, their experimental results are based on three different real data sets downloaded from the UCI repository. \vspace{-10pt}
\subsubsection{Parallel online sequential ELM: POS-ELM}\label{sec:pos-elm}
Wang et al. have been proposed parallel online sequential extreme learning machine (POS-ELM) method \cite{Wang2015224}. Main idea behind in this approach is to analyze the dependency relationships and the matrix calculations of OS-ELM \cite{4012031}. Their experimental results are based on nine different real data sets downloaded from the UCI repository.
\subsubsection{Distributed and Kernelized ELM: DK-ELM}
Bi et al. have been proposed both distributed and kernelized ELM (DK-ELM) based on MapReduce \cite{Bi2015456}. The difference between ELM and Kernelized ELM is that K-ELM applies kernels opposite to create random feature mappings. They provide a distributed implementation RBF kernel matrix calculation in massive data learning applications. Their experimental results are based on four different real data sets downloaded from the UCI repository and four synthetic data sets.
\subsubsection{ELM-MapReduce}\label{sec:elm-mr}
Chen et al. have been proposed MapReduce based ELM ensemble classifier called ELM-MapReduce, for large scale land cover classification of remote sensing data \cite{6565081}. Their approach contains two sequential phases: parallel training of multiple ELM classifiers and voting mechanism. In parallel training phase of proposed method, each $Map$ function computes an ELM classifier with a given training data set. In second phase called voting mechanism, a new MapReduce job is executed with a new partitioned test set into each $Map$ function with notation $data_j$. In $Reduce$ function of this phase, each $data_j$ is predicted with each ELM classifier trained in parallel training phase. Final classification predictions are the output of final $Reduce$ function. Therefore, this approach has a high communication cost. Their experimental results are based synthetic remote sensing image of training data.
\subsection{The Differences Between Proposed Model and Literature Review}
The main differences are:
\begin{itemize}
	\item In ELM $\star $, they use matrix multiplication decomposition. Each $Map$ function is responsible to calculate the Moore-Penrose generalized inverse operation. And their method produces one single classifier. In the proposed model in our paper, each $Reduce$ function produces ensemble classifier based on AdaBoost method. The final output ensemble classifier is a voting based combination of ensemble classifier trained in each $Reduce$ phase. 
	\item In OS-ELM based classification in hierarchical P2P Network, POS-ELM and DK-ELM, they propose ensemble classifier that combines multiple classifier trained with data chunks. Each peer classifier is learned from the local data. Therefore, each peer produces a single ELM classifier. In our method, each node (or peer) produces ensemble classifier to increase the classification accuracy.
	\item In ELM-MapReduce, they propose ensemble classifier with two different MapReduce jobs. In first MpaReduce job, their approach produces a single ELM classifier in each $Map$ function. In second MapReduce job, the test set is partitioned into each $Map$ function and produces final predicted labels based on the voting mechanism of ELM classifiers that are trained in the first MapReduce job. In our method, prediction is not included, our aim is to create a final ensemble classifier in only one MapReduce job.
\end{itemize}

Table \ref{tbl:litcompare} shows the main differences of all proposed methods. There are five different columns that are \textit{ensemble methods, single pass MapReduce, matrix multiplication, entire data set} and \textit{network communication}. \textit{Ensemble} column shows that the method builds a set of classifier function (i.e. ensemble model) to improve the accuracy performance of the final classification model. If an ensemble method is applied, then the performance of final model will have better accuracy result \cite{kuncheva2003measures}. \textit{Single Pass MapReduce} column shows that an iterative approach is not applied to the model. Entire learning phase is performed in a single pass of data through the job. \textit{Matrix Multiplication} column shows the \textit{hidden layer matrix} is calculated in each $Map$ function. The hidden layer matrix computation is a compute intensive operation. \textit{Entire Data Set}  column shows each $Map$ operation needs entire data set to build a final classifier model. \textit{Network Communication} column shows that each $MapReduce$ job needs to communicate with another job. Network communication will affect negatively on training time of the algorithm.

\begin{table}[h]
	\caption{The Differences Between Proposed Model and Literature Review.}
	\label{tbl:litcompare}
	\small
	\begin{center}
		\begin{tabular}{|c||c|p{1.5cm}|p{1.5cm}|p{1.5cm}|p{1.5cm}|}
			\hline Method & Ensemble & Single Pass MapReduce & Matrix Multiplication & Entire Data Set & Network Communication \\ 
			\hline \hline ELM $\star $ & No & Yes & No & Yes & No \\ 
			\hline OS-ELM & Yes & Yes & No & No & Yes \\
			\hline POS-ELM & Yes & Yes & No & Yes & No \\
			\hline DK-ELM & Yes & Yes & No & Yes & No \\
			\hline ELM-MapReduce & Yes & No & No & Yes & Yes \\
			\hline \hline Proposed Method & Yes & Yes & No & No & No \\
			\hline
		\end{tabular}
	\end{center}
\end{table}

\section{Preliminaries}\label{sec:preliminaries}
In this section, we introduce preliminary knowledge of ELM, AdaBoost and MapReduce briefly. 
\subsection{Extreme learning machine}\label{sec:ELM}
ELM was originally proposed for the single-hidden layer feedforward neural networks \cite{Huang06extremelearning,1650244,Huang2006489} . Then, ELM was extended to the generalized single-hidden layer feedforward networks where the hidden layer may not be neuron like \cite{Huang20073056,Huang20083460}. The main advantages of the ELM classification algorithm are that ELM can be trained hundred times faster than traditional neural network or support vector machine algorithm since its input weights and hidden node biases are randomly created and output layer weights can be analytically calculated by using a least-squares method \cite{6866146,Huang2008576}. The most noticeable feature of ELM is that its hidden layer parameters are selected randomly.

Given a set of training data $\mathcal{D}=\{(\mathbf{x}_i, y_i)\mid i=1,...,n\},\mathbf{x}_i \in \mathbb{R}^p,\, y_i \in \{1, 2,...,K\}\}$ sampled independently and identically distributed (i.i.d.) from some unknown distribution. The goal of a neural network is to learn a function $f:\mathcal{X} \rightarrow \mathcal{Y}$ where $\mathcal{X}$ is instance and $\mathcal{Y}$ is the set of all possible labels. The output label of an single hidden-layer feedforward neural networks (SLFNs) with $N$ hidden nodes can be described as
\begin{equation}
\label{eq:slfns}
f_N(\mathbf{x}) = \sum_{i=1}^{N}\beta_iG(\mathbf{a}_i,b_i,\mathbf{x}) , \, \mathbf{x} \in \mathbb{R}^n, \, \mathbf{a}_i \in \mathbb{R}^n
\end{equation}
where $\mathbf{a}_i$ and $b_i$ are the learning parameters of hidden nodes and $\beta_i$ is the weight connecting the $i$th hidden node to the output node.

The output function of ELM for generalized SLFNs can be identified by
\begin{equation}
\label{eq:slfnsgen}
f_N(\mathbf{x}) = \sum_{i=1}^{N}\beta_iG(\mathbf{a}_i,b_i,\mathbf{x}) = \mathbf{\beta} \times h(\mathbf{x})
\end{equation}

For the binary classification applications, the decision function of ELM becomes
\begin{equation}
\label{eq:binaryelm}
f_N(\mathbf{x}) = sign\left( \sum_{i=1}^{N}\beta_iG(\mathbf{a}_i,b_i,\mathbf{x}) \right) = sign\left(\mathbf{\beta} \times h(\mathbf{x}) \right)
\end{equation}

Equation \ref{eq:slfnsgen} can be written in another form as 
\begin{equation}
\label{eq:elm}
H\beta=T
\end{equation}
$H$ and $T$ are respectively hidden layer matrix and output matrix. Hidden layer matrix can be described as

\begin{equation}
\label{eq:H}
H(\tilde{a},\tilde{b},\tilde{x})= \begin{bmatrix} G(a_1,b_1,x_1) & \cdots & G(a_L,b_L,x_1) \\ \vdots & \ddots & \vdots \\ G(a_1,b_1,x_N) & \cdots & G(a_L,b_L,x_N) \end{bmatrix}_{N \times L}
\end{equation}
where $\tilde{a}=a_1,...,a_L$, $\tilde{b}=b_1,...,b_L$, $\tilde{x}=x_1,...,x_N$. Output matrix can be described as

\begin{equation}
\label{eq:elmoutput}
T= \begin{bmatrix} t_1  \hdots  t_N \end{bmatrix}^T
\end{equation}
The hidden nodes of SLFNs can be randomly generated. They can be independent of the training data.

\subsection{AdaBoost}\label{sec:AdaBoost}
The AdaBoost \cite{freund1995desicion} is a supervised learning algorithm designed to solve classification problems \cite{freund1999short}. The algorithm takes as input a training set $(\mathbf{x}_1, y_1),...,(\mathbf{x}_n, y_n)$ where the input sample $\mathbf{x}_i \in \mathbb{R}^p$, and the output value, $y_i$, in a finite space $y\in {1,...K}$. AdaBoost algorithm assumes, like ELM, a set of training data sampled independently and identically distributed (i.i.d.) from some unknown distribution $\mathcal{X}$. 

Given a space of feature vectors $X$ and two possible class labels, $y \in \{-1,+1\}$, AdaBoost goal is to learn a strong classifier $H(\mathbf{x})$ as a weighted ensemble of weak classifiers $h_t(\mathbf{x})$ predicting the label of any instance $\mathbf{x} \in X$ \cite{LandesaVazquez2013101}.
\begin{equation}
\label{eq:adaboost}
H(\mathbf{x}) = sign(f(\mathbf{x}))=sign\left(\sum_{t=1}^{T}\alpha_t h_t(\mathbf{x}) \right)
\end{equation}
Pseudocode for AdaBoost is given in Alg. \ref{alg:adaboost}. 
\begin{algorithm}
   \caption{The AdaBoost algorithm.}\label{alg:adaboost}
	\begin{algorithmic}[1]
		\Inputs{$\mathcal{D}=\{\{(\mathbf{x}_i, y_i)\mid i=1,...,n\},\mathbf{x}_i \in \mathbb{R}^p,\, y_i \in \{-1, +1\}\}_{i=1}^m$}
		\State Initialize $\mathcal{D}(i)=\frac{1}{m}$ for all $i$
		\While {$\,t<T$}
		\State Train \textbf{WeakLearner} using distribution $\mathcal{D}_t$
		\State get back a weak hypothesis $h_t : X \rightarrow \{1, 2,...,K\} $
		\State calculate the error of $h_t : \epsilon_t = Pr_{i \sim \mathcal{D}_t}[h_t(\mathbf{x}_i)\neq y_i]$
		\State Sets $\alpha_t =  \frac{1}{2}\ln{\frac{1-\epsilon_t}{\epsilon_t}}$
		\State update distribution $\mathcal{D}_{t+1} = \frac{\mathcal{D}_t}{Z_t} \times \begin{cases}
		e^{-\alpha_t}, & \text{if } h_t = y_i\\
		e^{\alpha_t},  & \text{if } h_t \neq y_i
		\end{cases}$ 
		\State equivalently $\mathcal{D}_{t+1} = \frac{\mathcal{D}_t \times exp(-\alpha_ty_ih_t(x_i))}{Z_t}$ where $Z_t$ is a normalization constant.
		\EndWhile
		\Outputs{final hypothesis $h^* = sign\left(\sum_{t=1}^{T}{\alpha_t h_t(\mathbf{x}) }\right) $}.
	\end{algorithmic}
\end{algorithm}
\subsection{MapReduce}\label{sec:MapReduce}
MapReduce is a new programming model to run parallel applications for large scale data sets processing to support data-intensive applications. It is derived from the map and reduce function combination from functional programming. Users specify a map function that processes a key/value pair to generate a set of intermediate key/value pairs, and a reduce function that merges all intermediate values associated with the same intermediate key. The MapReduce was originally developed by Google and built on principles in parallel manner \cite{Dean:2008:MSD:1327452.1327492}. The MapReduce framework first takes the input, divides it into smaller data chunks, and distributes them to worker nodes. MapReduce is divided into three major phases called map, reduce and a separated internal shuffle phase. The MapReduce framework automatically executes all those functions in a parallel manner over any number of processors/servers \cite{citeulike:4292765}.

Pseudo code of MapReduce framework is shown in Eq. \ref{eq:mapreduce}.
\begin{equation}
\label{eq:mapreduce}
\begin{split}
map(key_1,value_1) & \rightarrow list(key_2,value_2) \\
reduce(key_2,list(value_2)) & \rightarrow list(key_3,value_3)
\end{split}
\end{equation}

Mapreduce programming technique is widely used on different scientific fields, i.e. cyber-security \cite{soft01,soft02}, high energy physics \cite{bhimji2014hepdoop}, biology \cite{xu2014privacy}.
\section{Proposed Approach}\label{sec:approach}
In this section we provide the details of the MapReduce based distributed AdaBoosted ELM algorithm. The basic idea of AdaBoost-ELM based on MapReduce technique is introduced in Section \ref{sec:BasicIdea}. The MapReduce implementation of AdaBoosted ELM is described in Section \ref{sec:Implementation}.
\subsection{Basic Idea}\label{sec:BasicIdea}
Our main task is to parallel and distributed execute the computation of AdaBoosted ELM classification method. AdaBoosted ELM's basic idea is to calculate ensemble of classifier functions over partitioned data $(X_m,Y_m)$ in parallel manner. In Table \ref{tbl:notation}, a summary of commonly used  variables and notations to assess the classifier model performance of the AdaBoosted ELM method is given for convenience.
\begin{table}
	\caption{Commonly used variables and notations.}
	\label{tbl:notation}
	\small
	\begin{center}
		\begin{tabular}{|c||p{5cm}|}
			\hline Variables/Notation & Description \\ 
			\hline \hline $M$ & Data chunck split size \\ 
			$h$ & A single classifier function \\
			$X_m$ & Data chunck $m$ of \textit{input} values of $\mathcal{D}$ \\
			$Y_m$ & Data chunck $m$ of \textit{output} values of $\mathcal{D}$ \\
			$\epsilon$ & Error rate \\
			\# Chunk & Number of data chunk \\
			$T$ & AdaBoost $T$ size \\
			\# H. Nodes & Number of hidden nodes used in ELM \\
			Acc & Accuracy of classifier hypothesis \\
			$k$ & Number of classes \\
			\hline 
		\end{tabular} 
	\end{center}
\end{table}
\subsection{Analysis of the proposed algorithm}\label{sec:analysisofalg}
Barlett showed that  the size of the weights is more important than the size of the neural network \cite{661502}. Kragh et al. also showed that ensemble methods of neural networks get better accuracy performance over unseen examples \cite{Krogh95neuralnetwork}. The main motivation of the this work is the idea that small size ELM ensembles can obtain more accurate classifier model that are comparable to individual classifiers.

In the proposed model, at every data chunk, there is a set of classifier functions that acts as a single classification model. The single model at every data chunk $m$ is defined as follows:
\begin{equation}
\label{eq:setELM}
f^{(m)}(\mathbf{x}) = \argmax_k \sum_{t=1}^{T}{\alpha_t h_t(\mathbf{x})}
\end{equation}
The selected ensemble ELM classifier models from the reduce phase of MapReduce algorithm are combined into one single classification model.
\begin{equation}
\label{eq:setELMFinal}
\hat{h}(\mathbf{x}) = \argmax_k \sum_{i=1}^{m}{f^{(m)}(\mathbf{x})}
\end{equation}
\subsection{Implementation of the Model}\label{sec:Implementation}
The pseudocodes of MapReduce-based AdaBoost ELM are shown in Algorithm \ref{alg:map} and Algorithm \ref{alg:reduce}. The $Map$ procedure of our training model is implemented based on random assignment of each row of the training data set with split size of data, $M$, in line 2 of Algorithm \ref{alg:map}. The input, $\mathbf{x}$ , is a row of traing data set $\mathcal{D}$. $Map$ procedure partition the input matrix by row, producing $<randomSplitId,\mathbf{x}>$ key-value pairs. $randomSplitId$ is the identifier of the data chunk and is transferred as the input key to $Reduce$ phase.
\begin{algorithm}
	\caption{AdaBoostedELM::Map}\label{alg:map}
	\begin{algorithmic}[1]
		\Inputs{Training record $(\mathbf{x},y) \in \mathcal{D}$, Data set split size $M$}
		\State $k \gets rand(0,M)$
		\State $Output(k,(\mathbf{x},y))$
	\end{algorithmic}
\end{algorithm}
The pseudo code of $Reduce$ phase is shown in Algorithm \ref{alg:reduce}. $Reduce$ procedure is implemented based on the for-loop of lines 3 -– 8 of Algorithm \ref{alg:reduce}. The output ELM classifier of sub data set $(\mathbf{X}_k,\mathbf{y}_k)$ is calculated using AdaBoost constantly block by block, so every reduce task completes training phase and outputs an AdaBoosted set of classifier functions. The $mapper$'s input $k$ is the $randomSplitId$ to create the data chunk and created in the $Map$ phase of our training model. 
\begin{algorithm}
	\caption{AdaBoostedELM::Reduce}\label{alg:reduce}
	\begin{algorithmic}[1]
		\Inputs{Key $k$, Value Set $V$,AdaBoost Iteration Size $T$}
		\State Split $V$ into input space $\mathbf{X}_n$ and out labels $\mathbf{y_n}$ with $(\mathbf{X}_n,\mathbf{y}_n) \gets V$
		\For{$t=1..T$}
		\State Train sub data set with ELM: $h_t \gets ELM(\mathbf{X},y)$
		\State $\mathbf{y}_{pred},\epsilon_t \gets h_t(X)$
		\State $\alpha_t \gets  \frac{1}{2}\ln{\frac{1-\epsilon_t}{\epsilon_t}}$
		\State $\mathcal{D}_{t+1} = \frac{\mathcal{D}_t \times exp(-\alpha_ty_ih_t(x_i))}{Z_t}$
		\EndFor
		\Outputs{Final hypothesis for the reduce function $m$ : $h_m \gets \argmax_k \sum_{t=1}^{T}{\alpha_t h_t(\mathbf{x}) } $}.
	\end{algorithmic}
\end{algorithm}
\section{Experiments}\label{sec:experiments}
In this section, we perform experiments on real-world data sets from the public available data set repositories. Public data sets are used to evaluate the proposed learning method. Then, classification models of each data set are compared for accuracy results  with the single instance of learning algorithm performance. 

In Section \ref{sec:expsetup} we explain the data sets  and parameters that are used in experiments. The conventional ELM is applied all data sets and we find the accuracy performance over number of hidden nodes in Section \ref{sec:conv_elm}. In Section \ref{sec:eval}, we show the empirical results of proposed distributed adaboost ELM training algorithm.
\subsection{Experimental setup}\label{sec:expsetup}
In this section we apply our approach to five different data sets to verify its effectivity and efficiency. To demonstrate the effectiveness and performance of the proposed model, we apply it on various classification data sets from public data set repositories. To obtain an optimal value of Mapper size, $m$, we range it in the range from 20 to 100.
\subsubsection{Commonly Used Classification Data Sets}\label{sec:commonds}
We experiment on five public data sets which are summarized in Table \ref{tbl:dslist}, including Pendigit, Letter, Statlog, Page-blocks and Waveform. They are all multiclass data sets. All experiments are repeated 5 times and the results are averaged. All data sets are publicly available in svmlight format on the LIBSVM web site \cite{libsvmdatasets}.

\textit{Pendigit} data set is a collection of pen-based recognition of handwritten digits \cite{Alimoglu96methodsof}. The data set contains 250 samples from 44 people. The first 7494 instances written by 30 people are used for the training data set, and the digits written by other 14 people are used for the independent testing purpose. 

\textit{Skin} data set is a collection of skin segmentation constructed over R, G, B color space \cite{5409447}. The data set contains face images of different age groups (young, middle, old), genders and racial groups (White, Black, Asian). The data set contains 245057 instances; out of which 50859 is the skin labeled instances and 194198 is non-skin instances. 

\textit{Statlog / Shuttle} data set is a collection of space shuttle created by NASA \cite{Hsu:2002:CMM:2325784.2326870}. The data set contains 43500 training instances and 14500 testing instances.  80\% of the data belongs to class 1.

\textit{Page Blocks} data set is a collection of page layout of a document that has been detected by a segmentation process \cite{Malerba96afurther}. The data set contains 4500 training instances and 973 testing instances.

\textit{Waveform} data set is a collection of Breiman's waveform domains of CART book's \cite{breiman1984classification}. The data set contains 4400 training instances and 600 testing instances.

\begin{table}[h]
	\caption{Description of the testing data sets used in the experiments.}
	\label{tbl:dslist}
	\begin{center}
		\begin{tabular}{|c||r|r|c|c|}
			\hline Data set & \#Train & \#Test & \#Classes & \#Attributes \\ 
			\hline \hline Pendigit & 7494 & 3498 & 10 & 16 \\ 
			\hline Skin & 220543 & 24507 & 2 & 3 \\ 
			\hline Statlog / Shuttle & 43500 & 14500 & 7 & 9 \\ 
			\hline Page-blocks & 4500 & 973 & 5 & 10 \\
			\hline Waveform & 4400 & 600 & 3 & 21 \\
			\hline 
		\end{tabular} 
	\end{center}
\end{table}

\subsubsection{Large Scale Classification Data Sets}\label{sec:commonlargeds}
We experiment on three public large scale data sets which are summarized in Table \ref{tbl:dslistlarge}, including "\textit{Record Linkage Comparison Patterns (Donation) }", "\textit{SUSY}" and "\textit{HIGGS}". All experiments are repeated 5 times and the results are averaged. 

\textit{Donation} represent individual data, including first and family name, sex, date of birth and postal code, which were collected through iterative insertions in the course of several years. The comparison patterns in this data set are based on a sample of 100.000 records dating from 2005 to 2008 \cite{schmidtmann2009evaluation}. The data set contains 5,749,132 training instances and 1,000,000 testing instances. The data set is available on UCI web site \cite{ucidonation}.

\textit{SUSY} is a classification data set that distinguish between a signal process which produces supersymmetric particles and a background process which does not \cite{baldi2014searching}. The first 8 features are kinematic properties measured by the particle detectors in the accelerator. The last ten features are functions of the first 8 features. The data set contains 5,000,000 training instances and 50,000 testing instances. The data set is available on UCI web site \cite{ucisusy}.

\textit{HIGSS} is a classification problem to distinguish between a signal process which produces Higgs bosons and a background process which does not \cite{baldi2014searching}. The first 21 features (columns 2-22) are kinematic properties measured by the particle detectors in the accelerator. The last seven features are functions of the first 21 features. The data set contains 11,000,000 training instances and 500,000 testing instances. The data set is available on UCI web site \cite{ucihiggs}.

\begin{table}[h]
	\caption{Description of the testing large scale data sets used in the experiments.}
	\label{tbl:dslistlarge}
	\begin{center}
		\begin{tabular}{|c||r|r|c|c|}
			\hline Data set & \#Train & \#Test & \#Classes & \#Attributes \\ 
			\hline \hline Donation & 5,749,132 & 1,000,000 & 2 & 12 \\ 
			\hline SUSY & 5,000,000 & 50,000 & 2 & 18 \\ 
			\hline HIGSS & 11,000,000 & 1,000,000 & 2 & 28 \\
			\hline 
		\end{tabular} 
	\end{center}
\end{table}

\subsection{Evaluation}\label{sec:eval}
Since the data sets that are used in our experiments are highly imbalanced, traditional accuracy based performance evaluation is not enough to find out an optimal classifier. We used four different metrics, the overall prediction accuracy, average recall, average precision \cite{Turpin:2006:UPV:1148170.1148176} and $F$-score, to evaluate the classification accuracy which are common measurement metrics in information retrieval \cite{Manning:2008:IIR:1394399,Makhoul99performancemeasures}. 

Precision is defined as the fraction of retrieved samples that are relevant. Precision is shown in Eq. \ref{eqn:prec}. 
\begin{equation}
\label{eqn:prec}
Precision = \frac{Correct}{Correct + False}
\end{equation}
Recall is defined as the fraction of relevant samples that is retrieved. Recall is shown in Eq. \ref{eqn:recall}.
\begin{equation}
\label{eqn:recall}
Precision = \frac{Correct}{Correct + Missed}
\end{equation}
The proposed evaluation model calculates the precision and recall for each class from prediction scores then finds their mean. Average precision and recall is shown in Eq. \ref{eqn:avgprec} and Eq. \ref{eqn:avgrecall}.
\begin{equation}
\label{eqn:avgprec}
Precision_{avg} = \frac{1}{n_{classes}}\sum_{i=0}^{n_{classes}-1}{Prec_i}
\end{equation}
\begin{equation}
\label{eqn:avgrecall}
Recall_{avg} = \frac{1}{n_{classes}}\sum_{i=0}^{n_{classes}-1}{Recall_i}
\end{equation}
$F$-measure is defined as the harmonic mean of precision and recall. The 
\begin{equation}
\label{eqn:fmeasure}
F_1 = 2 \times \frac{Prec_{avg} \times Recall_{avg}}{Prec_{avg} + Recall_{avg}}
\end{equation}

\subsection{Data set results with conventional ELM}\label{sec:conv_elm}
\begin{figure}
	\begin{subfigure}[b]{0.50\textwidth}
		\includegraphics[width=1\linewidth]{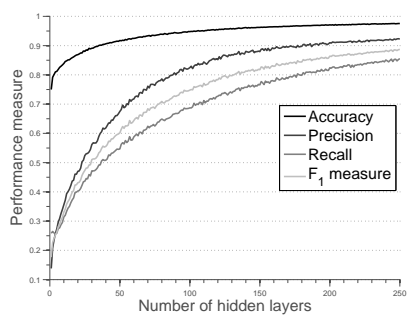}
		\caption{Statlog data set.}
		\label{fig:statlog_conv}
	\end{subfigure}
	\hspace{1pt}
	\begin{subfigure}[b]{0.50\textwidth}
		\includegraphics[width=1\linewidth]{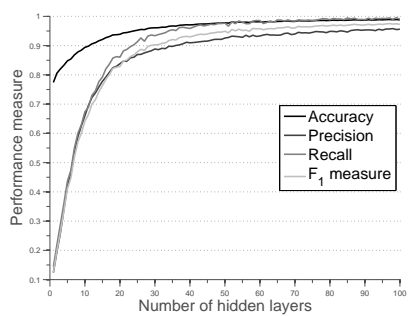}
		\caption{Skin data set.}
		\label{fig:skin_conv}
	\end{subfigure}
	\hspace{1pt}
	\begin{subfigure}[b]{0.50\textwidth}
		\includegraphics[width=1\linewidth]{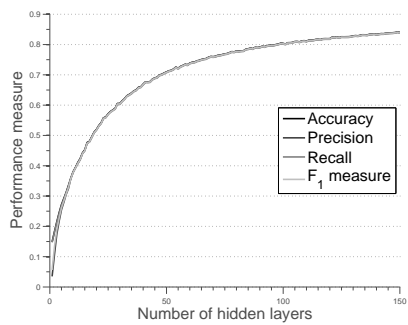}
		\caption{Pen digit data set.}
		\label{fig:pendigit_conv}
	\end{subfigure}
	\hspace{1pt}
	\begin{subfigure}[b]{0.50\textwidth}
		\includegraphics[width=1\linewidth]{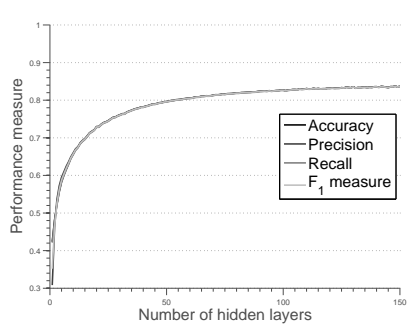}
		\caption{Waveform data set.}
		\label{fig:waveform_conv}
	\end{subfigure}
	\hspace{1pt}
	\begin{subfigure}[b]{0.50\textwidth}
		\includegraphics[width=1\linewidth]{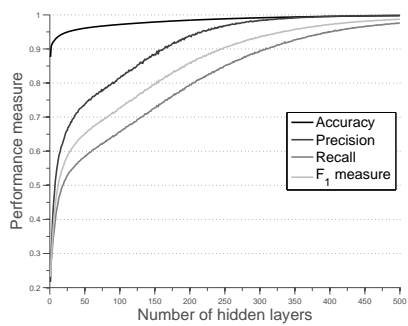}
		\caption{Page blocks data set.}
		\label{fig:page-blocks_conv}
	\end{subfigure}
	
	\caption{Number of hidden nodes in ELM versus classifier precision.}
	\label{fig:convelm}
\end{figure}

Figure \ref{fig:convelm} shows that the accuracy performance of ELM for experimental data sets becomes steady-state after a threshold value of $N$. The testing classification performance is measured through accuracy, precision, recall and $F_1$ measure. $N$ varies from 150 to 500. 

Table \ref{tbl:dsconvtab} shows the best performance of the conventional ELM method of each data set.
\begin{table}[h]
	\caption{Data set results with conventional ELM.}
	\label{tbl:dsconvtab}
	\begin{center}
		\begin{tabular}{|c||r|r|c|c|}
			\hline Data set & $F_1$ & Recall & Precision & Accuracy \\ 
			\hline \hline Pendigit &  0.8404 & 0.8393 & 0.8416 & 0.8407 \\ 
			\hline Skin & 0.9754 & 0.9956 & 0.9583 & 0.9894 \\ 
			\hline Statlog & 0.8871 & 0.8556 & 0.9237 & 0.9757 \\ 
			\hline Page-blocks & 0.9873 & 0.9764 & 0.9988 & 0.9977 \\
			\hline Waveform & 0.8372 & 0.8368 & 0.8375 & 0.8376 \\
			\hline 
		\end{tabular} 
	\end{center}
\end{table}

The conventional ELM training algorithm can be applied only in Section \ref{sec:commonds}. The large scale data sets in Section \ref{sec:commonlargeds} are not feasible to train on a single computer.

\subsection{Testing Accuracy Analysis}
Because of two different data set type ("commonly used", "large scale") are used, the results are divided into two different sections. In Section \ref{sec:cucds}, the figures and the plots show the implementation results of commonly used classification data sets. Section \ref{sec:lsdsacc} shows the large scale data sets results.
\subsubsection{Commonly Used Classification Data Sets}\label{sec:cucds}
\begin{figure}
	\begin{subfigure}[b]{0.32\textwidth}
		\includegraphics[width=1\linewidth]{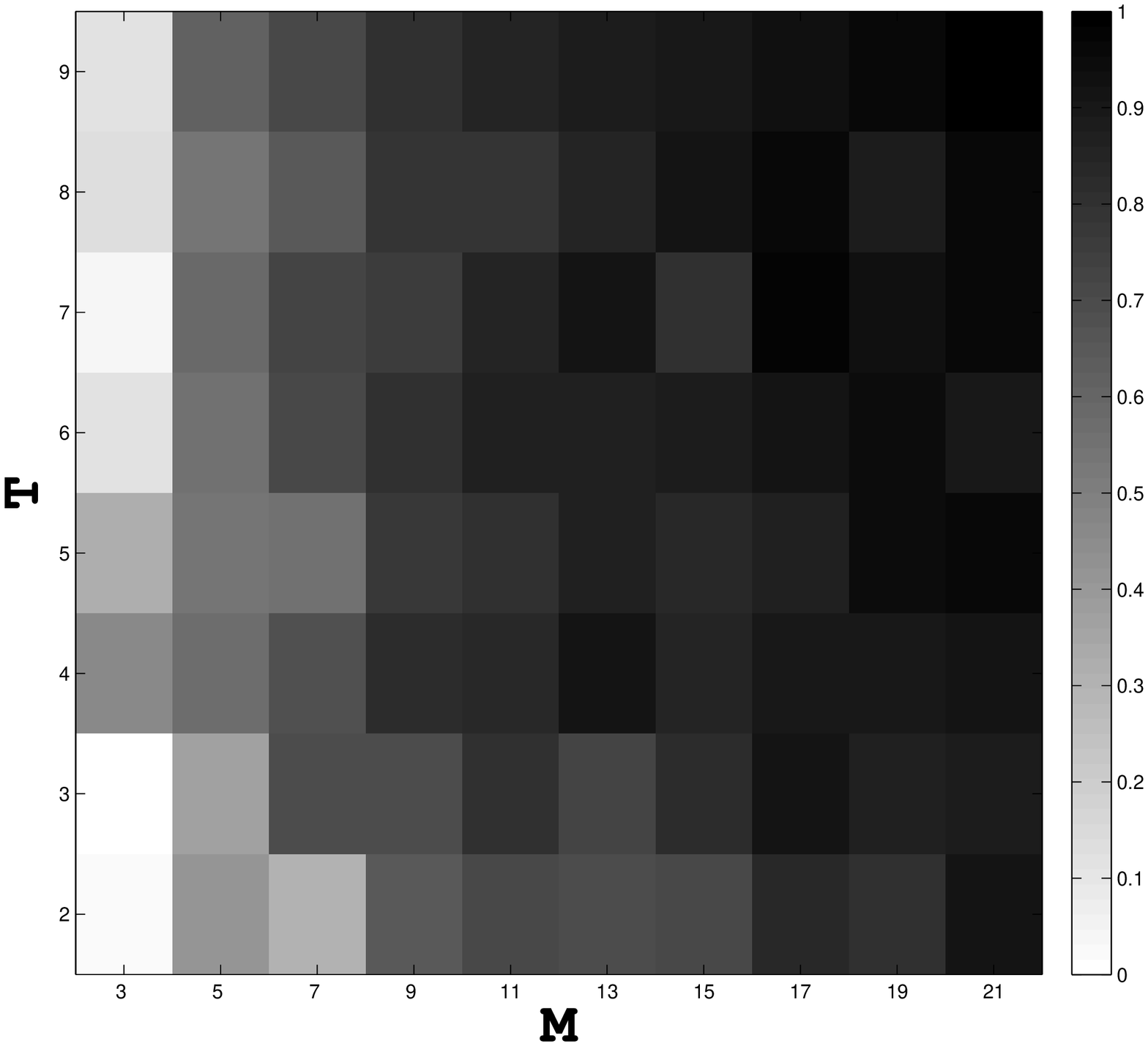}
		\caption{Split size and adaboost $T$ size}
		\label{fig:statlog_map_T}
	\end{subfigure}
	\begin{subfigure}[b]{0.32\textwidth}
		\includegraphics[width=1\linewidth]{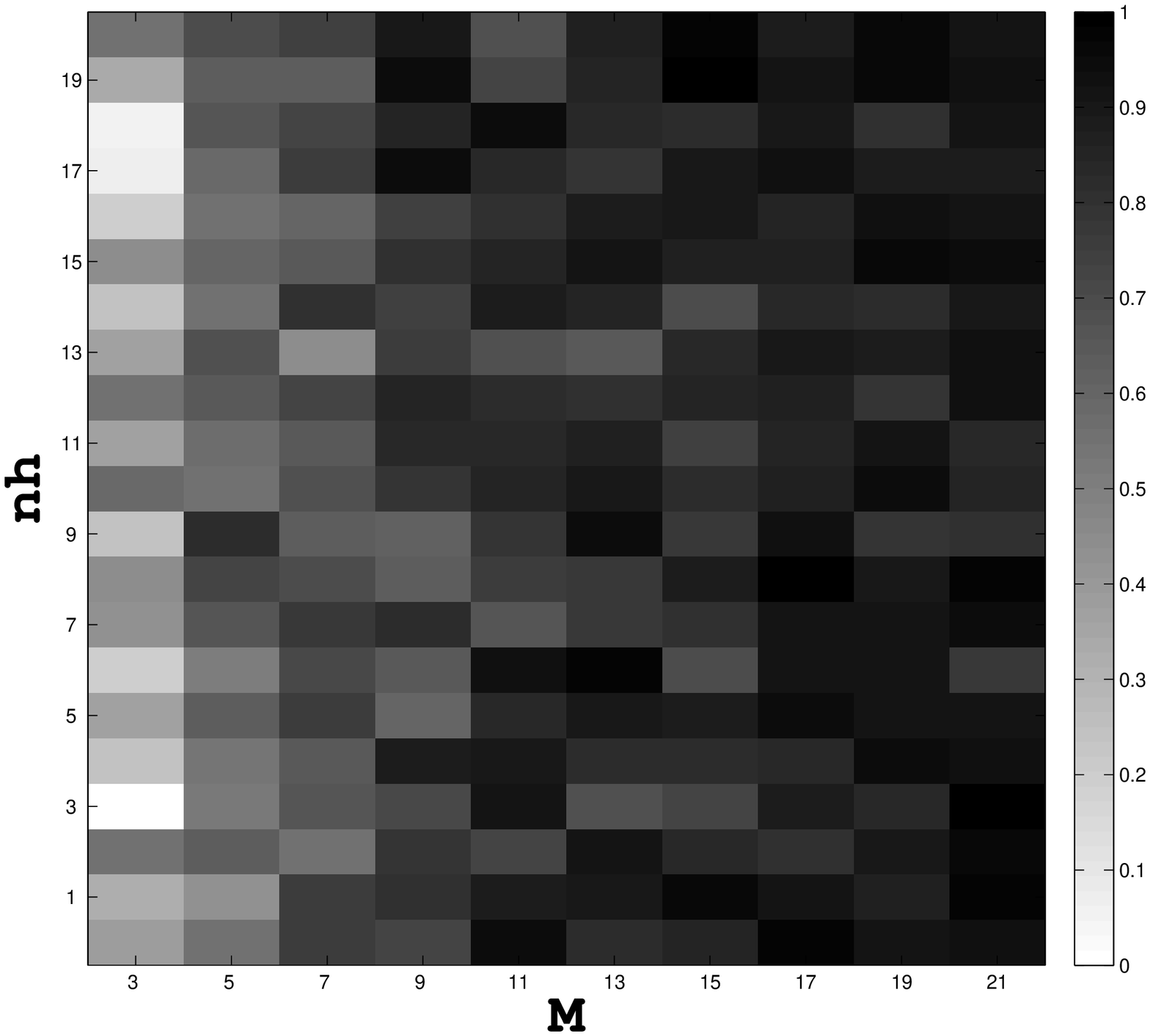}
		\caption{Split size and number of $nh$.}
		\label{fig:statlog_map_nh}
	\end{subfigure}
	\begin{subfigure}[b]{0.32\textwidth}
		\includegraphics[width=1\linewidth]{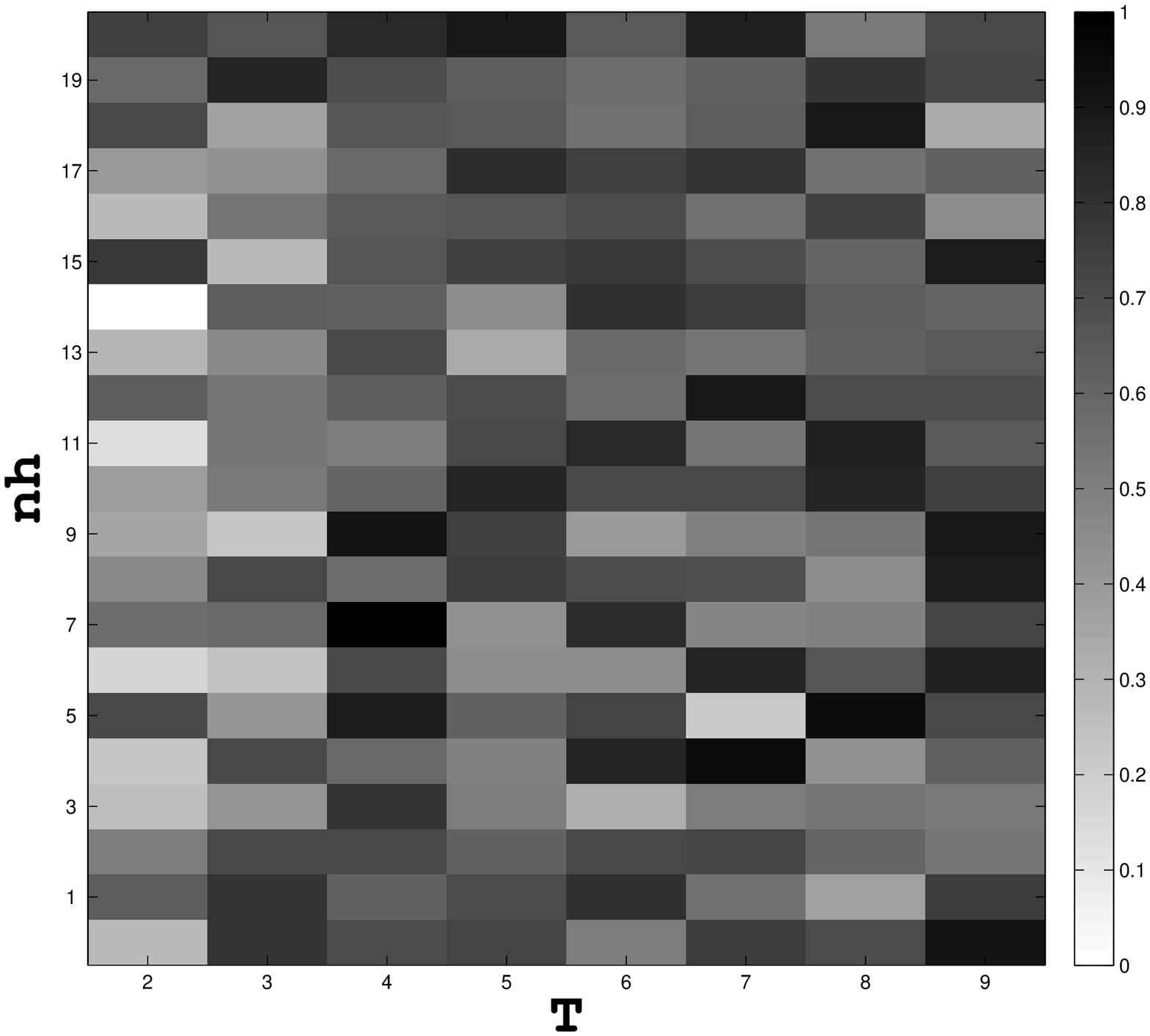}
		\caption{Adaboost $T$ size and number of $nh$.}
		\label{fig:statlog_T_nh}
	\end{subfigure}
	\caption{Statlog data set heatmap.}
	\label{fig:statlogres}
\end{figure}

\begin{figure}
	\begin{subfigure}[b]{0.32\textwidth}
		\includegraphics[width=1\linewidth]{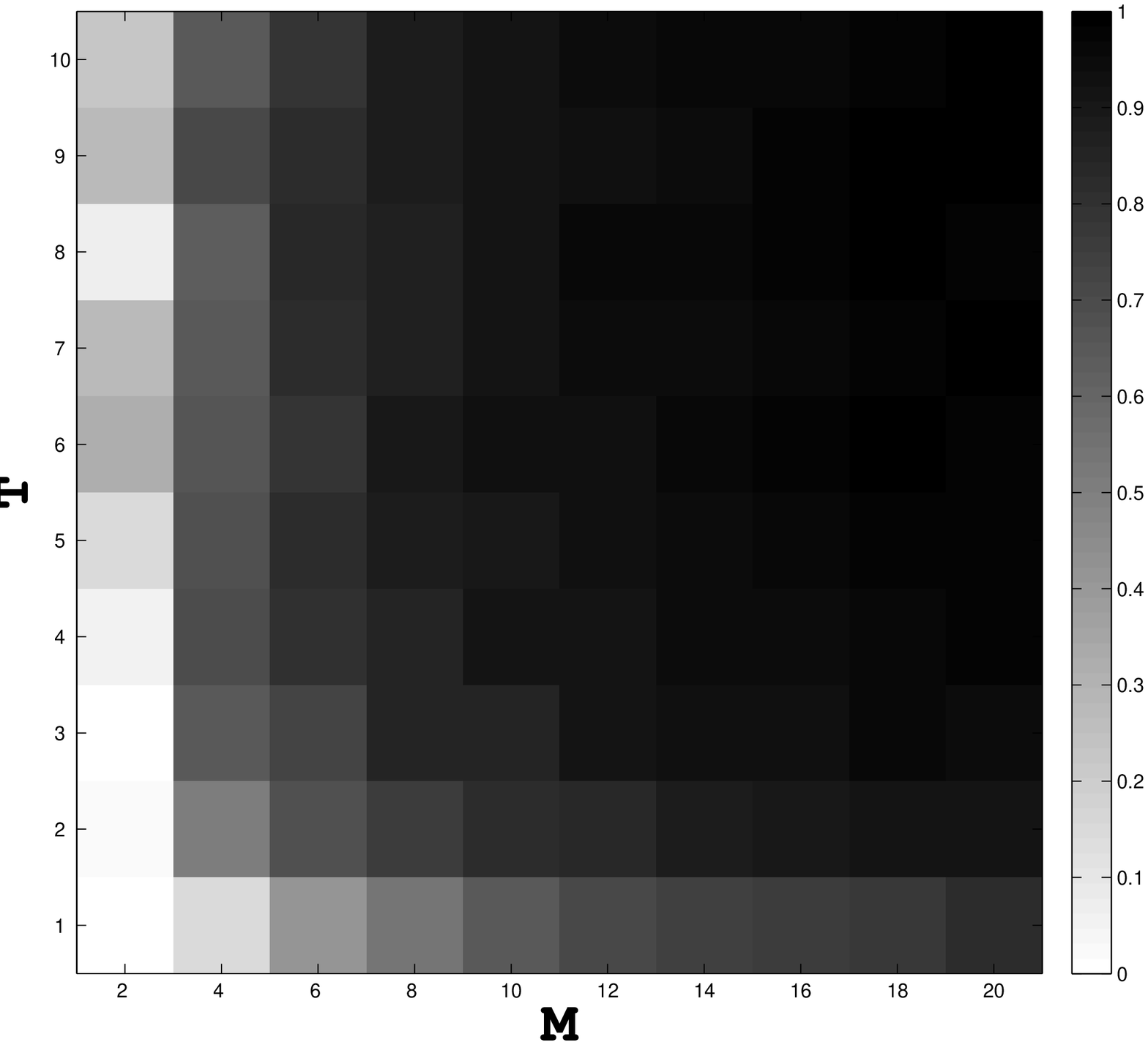}
		\caption{Split size and adaboost $T$ size}
		\label{fig:pendigit_map_T}
	\end{subfigure}
	\begin{subfigure}[b]{0.32\textwidth}
		\includegraphics[width=1\linewidth]{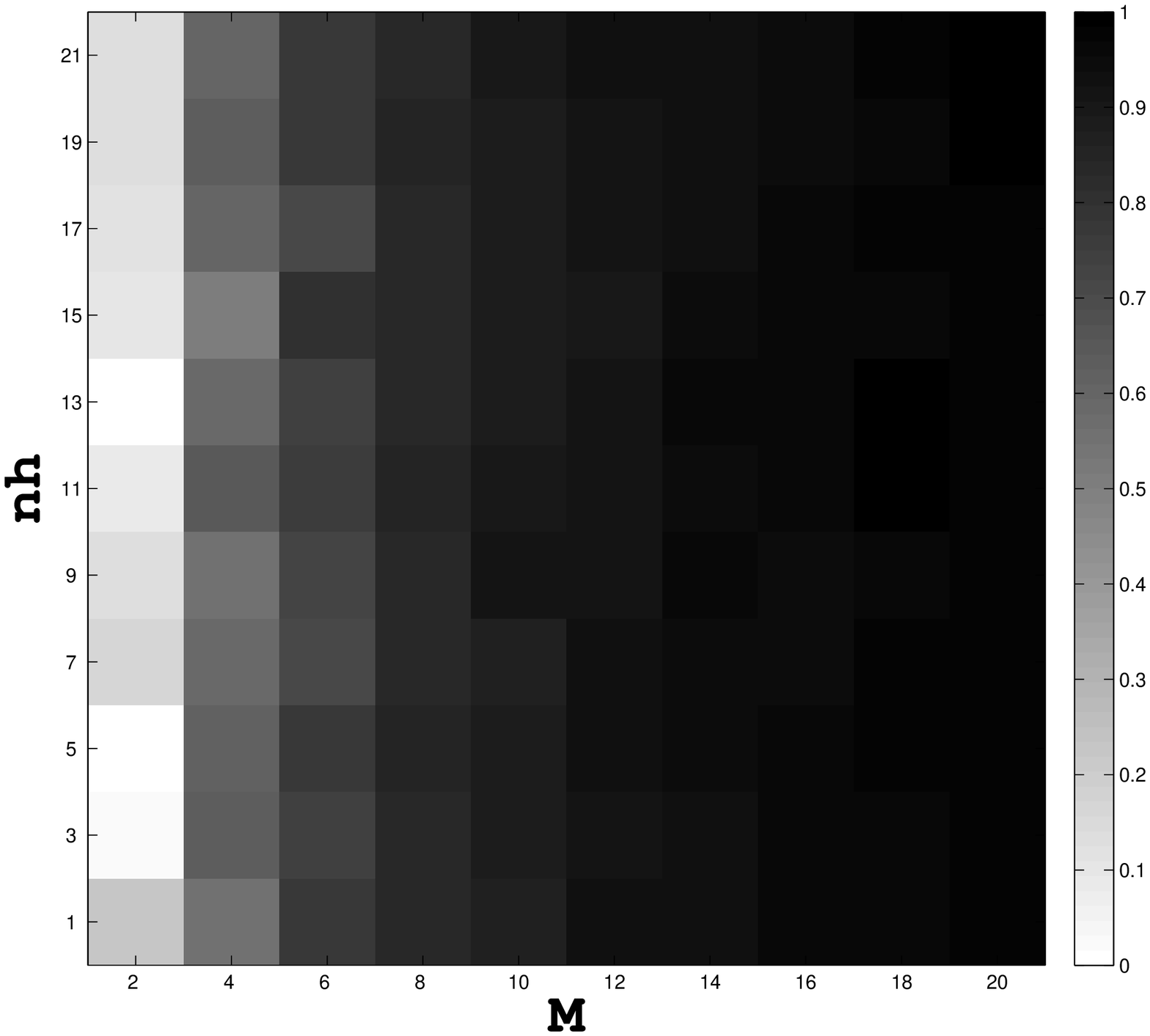}
		\caption{Split size and number of $nh$.}
		\label{fig:pendigit_map_nh}
	\end{subfigure}
	\begin{subfigure}[b]{0.32\textwidth}
		\includegraphics[width=1\linewidth]{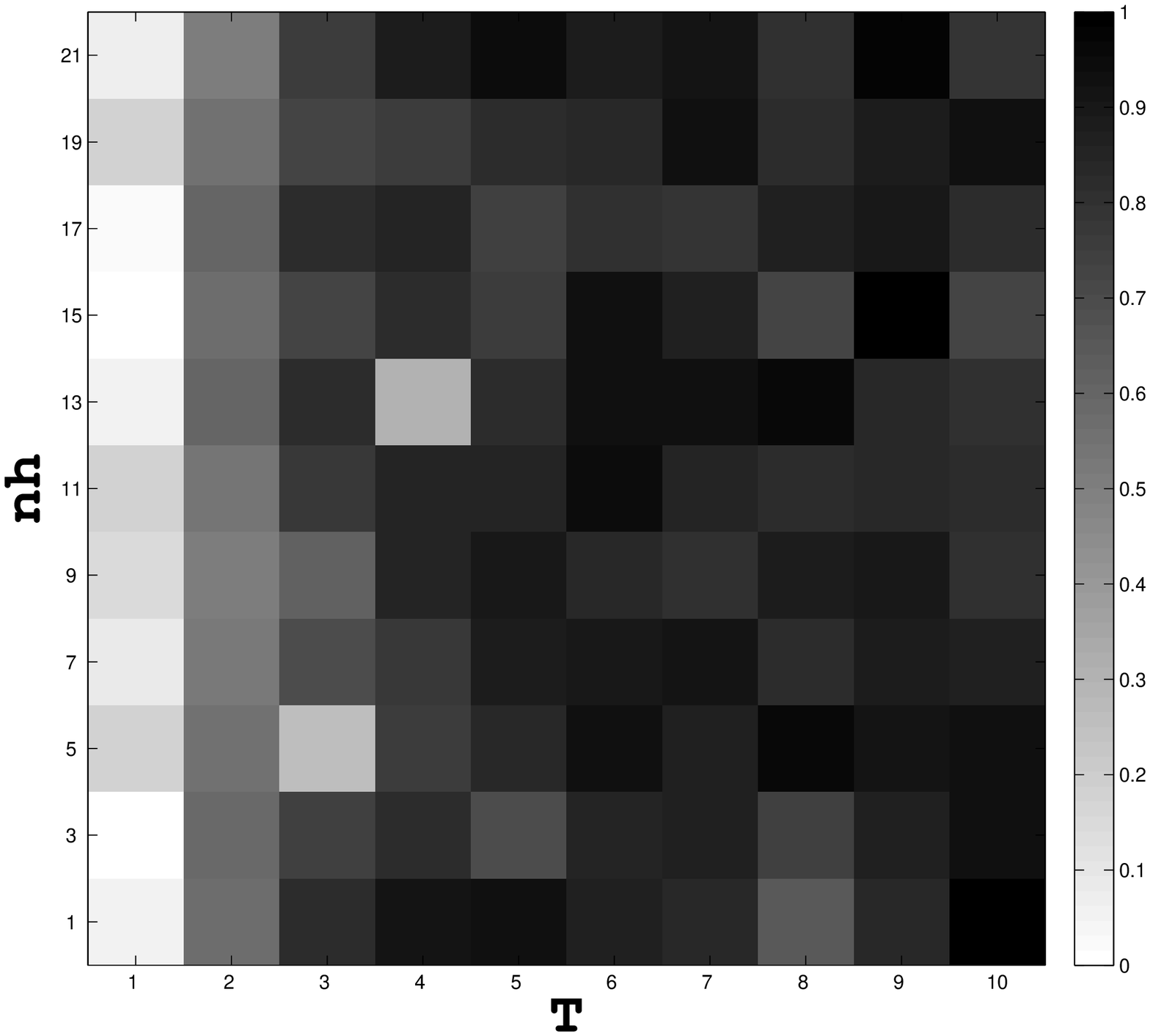}
		\caption{Adaboost $T$ size and number of $nh$.}
		\label{fig:pendigit_T_nh}
	\end{subfigure}
	\caption{Pendigit data set heatmap.}
	\label{fig:pendigitres}
\end{figure}

\begin{figure}
	\begin{subfigure}[b]{0.32\textwidth}
		\includegraphics[width=1\linewidth]{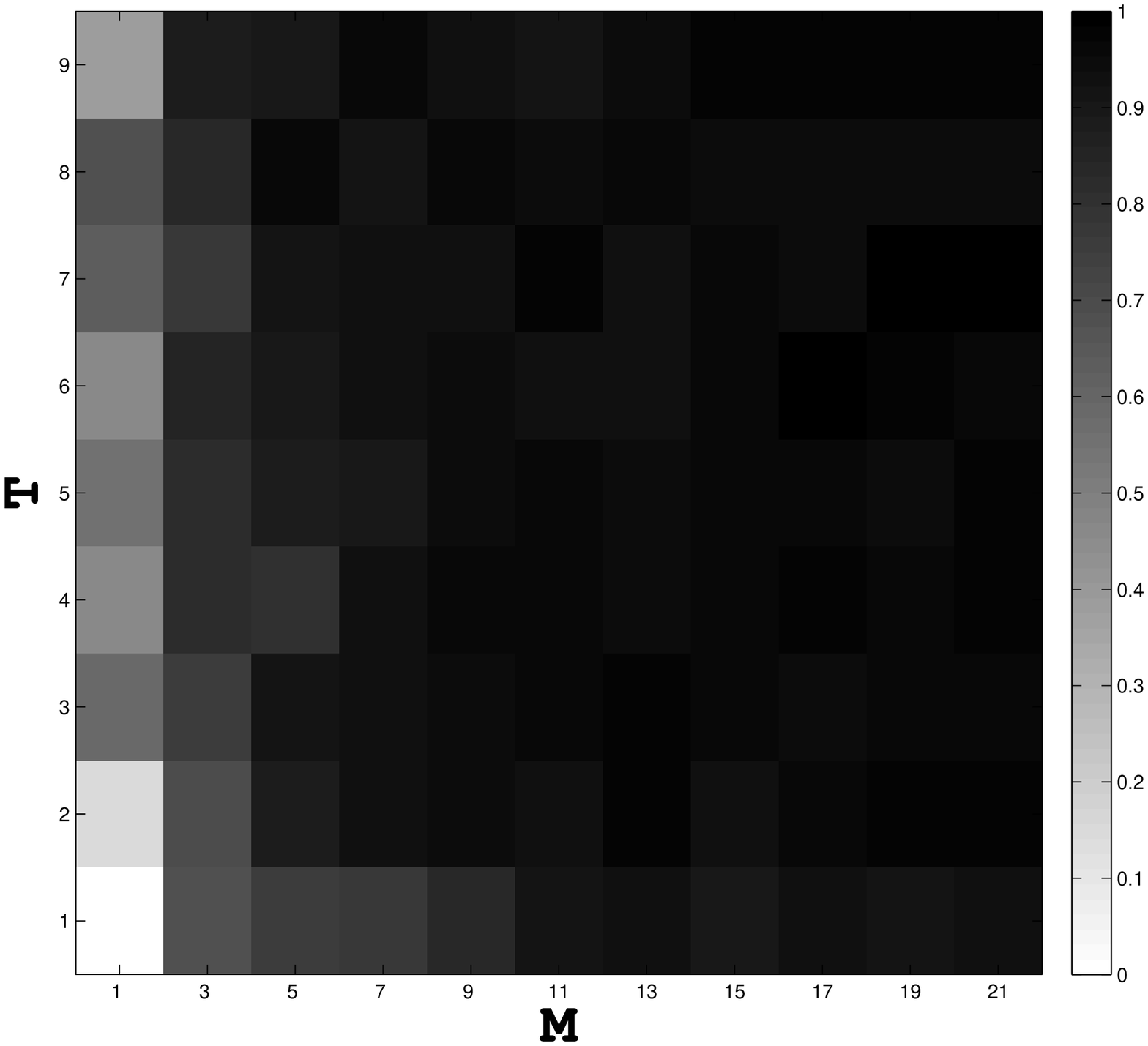}
		\caption{Split size and adaboost $T$ size}
		\label{fig:skin_map_T}
	\end{subfigure}
	\begin{subfigure}[b]{0.32\textwidth}
		\includegraphics[width=1\linewidth]{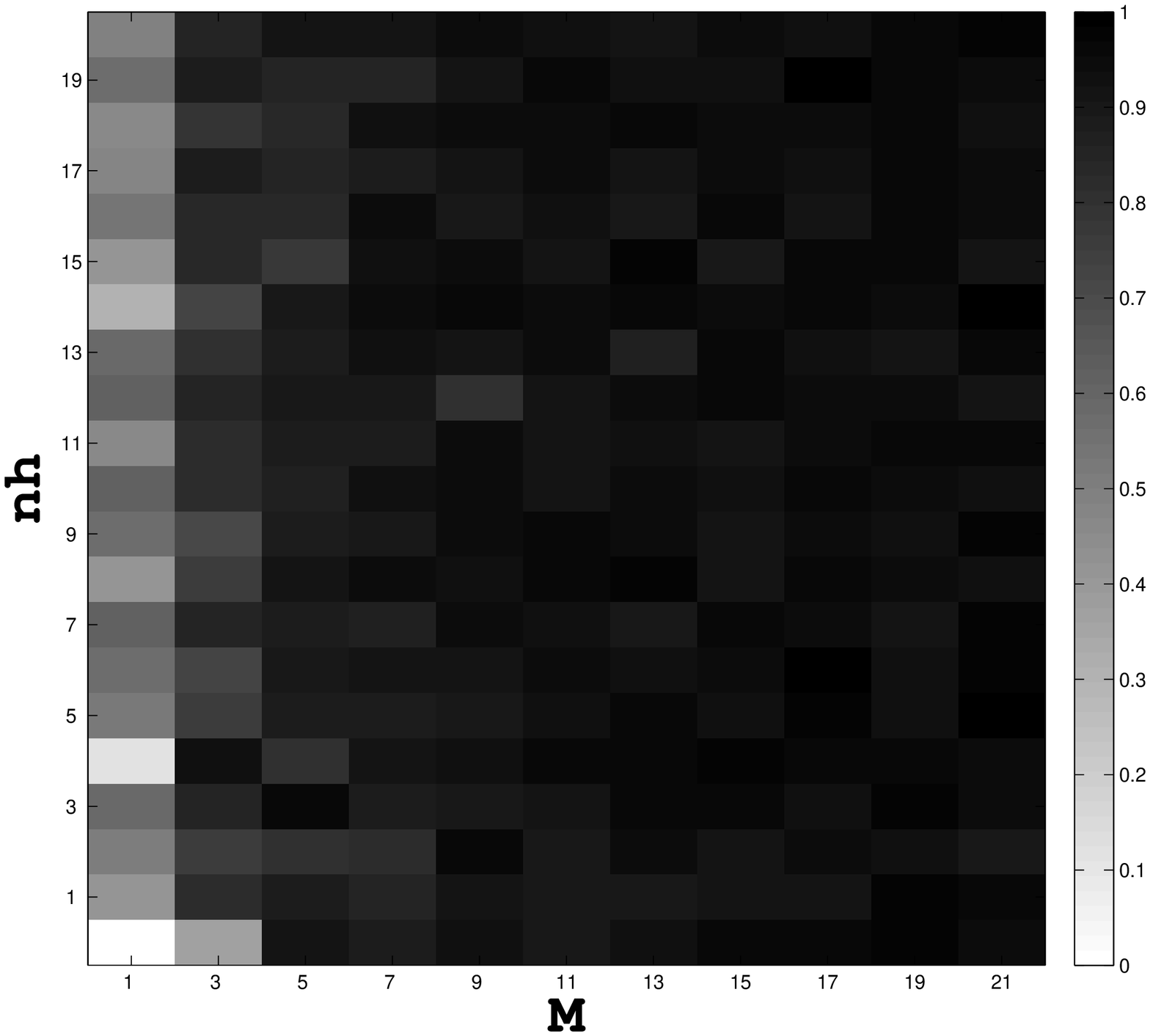}
		\caption{Split size and number of $nh$.}
		\label{fig:skin_map_nh}
	\end{subfigure}
	\begin{subfigure}[b]{0.32\textwidth}
		\includegraphics[width=1\linewidth]{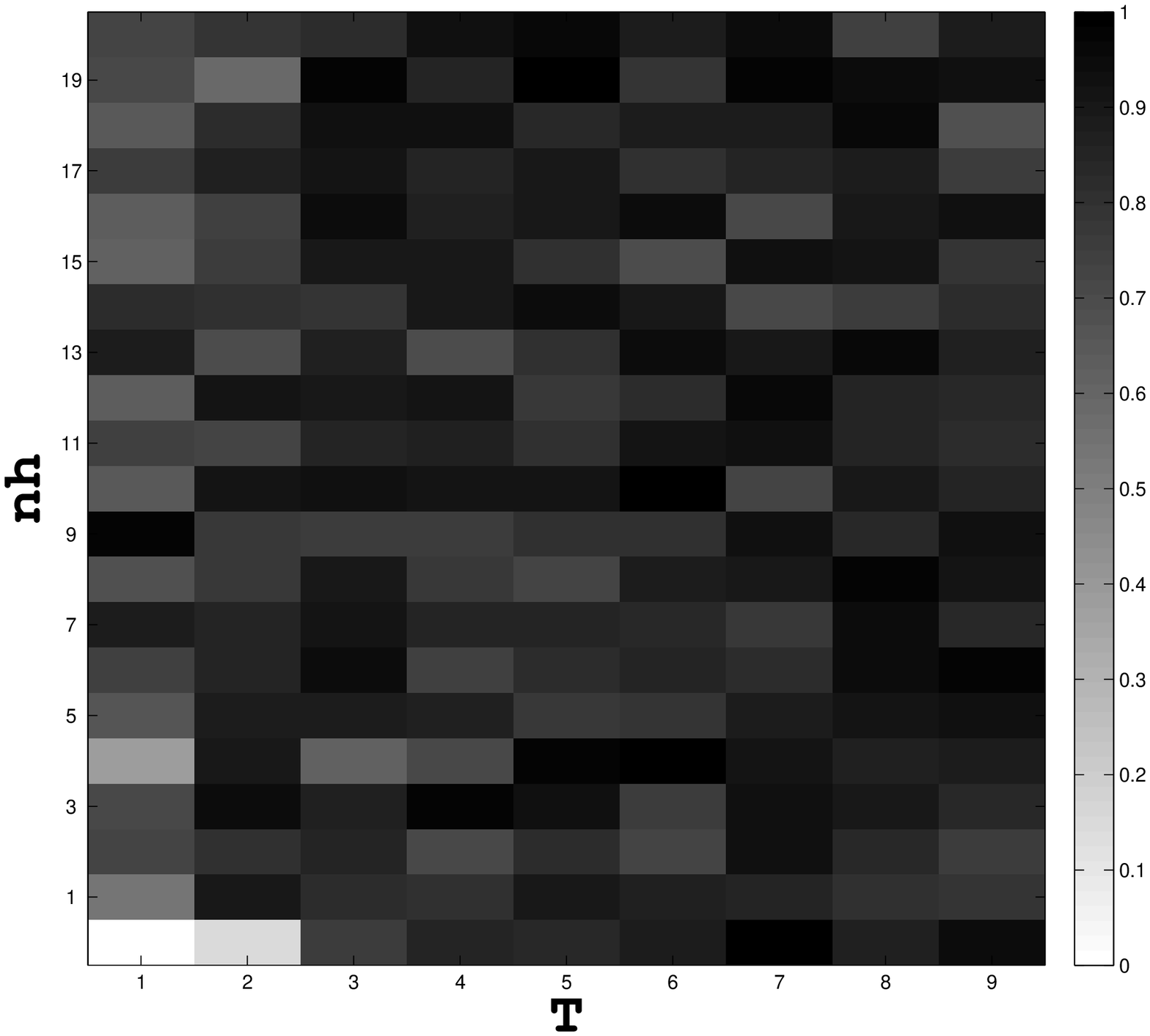}
		\caption{Adaboost $T$ size and number of $nh$.}
		\label{fig:skin_T_nh}
	\end{subfigure}
	\caption{Skin data set heatmap.}
	\label{fig:skinres}
\end{figure}

\begin{figure}
	\begin{subfigure}[b]{0.32\textwidth}
		\includegraphics[width=1\linewidth]{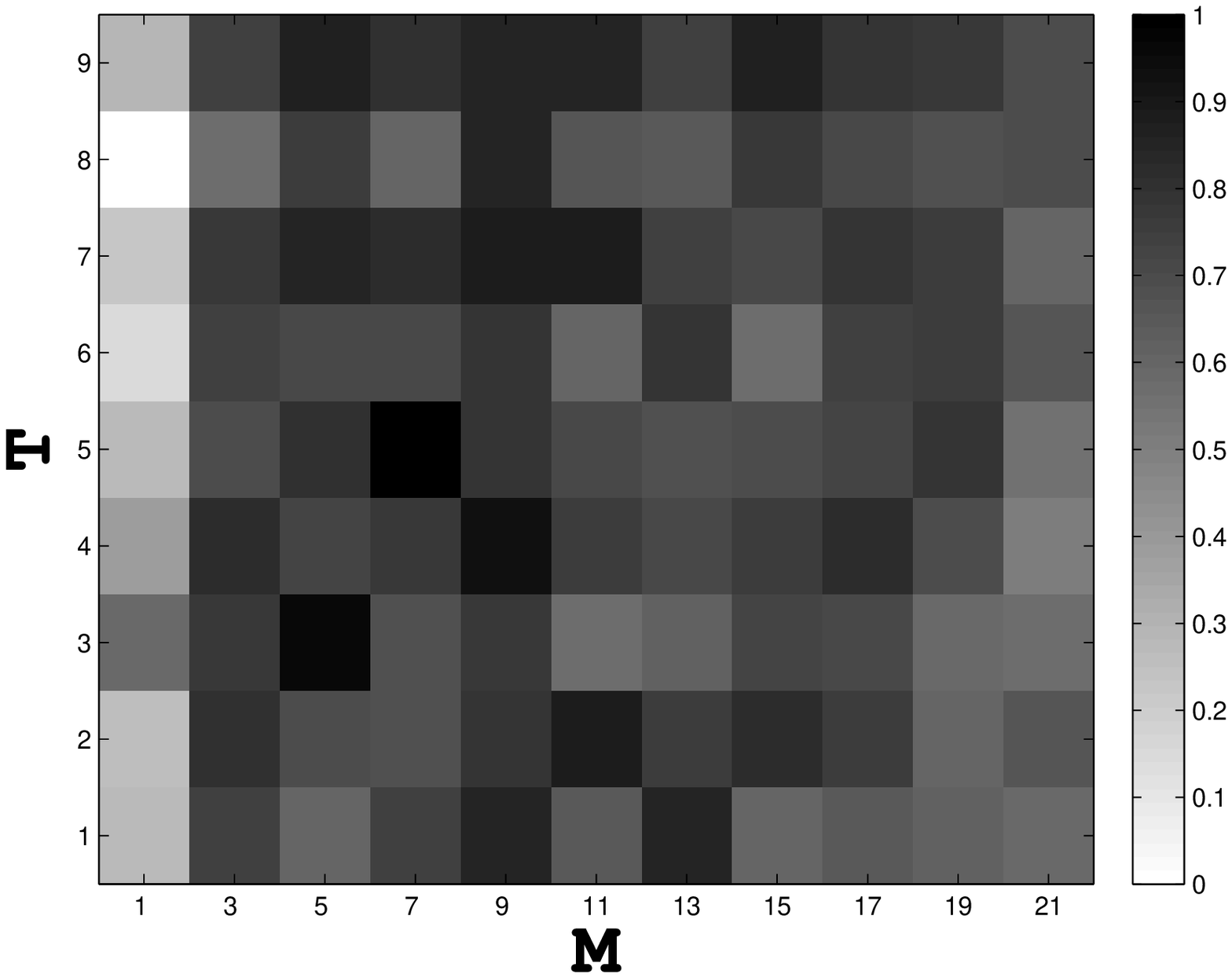}
		\caption{Split size and adaboost $T$ size}
		\label{fig:pageblocks_map_T}
	\end{subfigure}
	\begin{subfigure}[b]{0.32\textwidth}
		\includegraphics[width=1\linewidth]{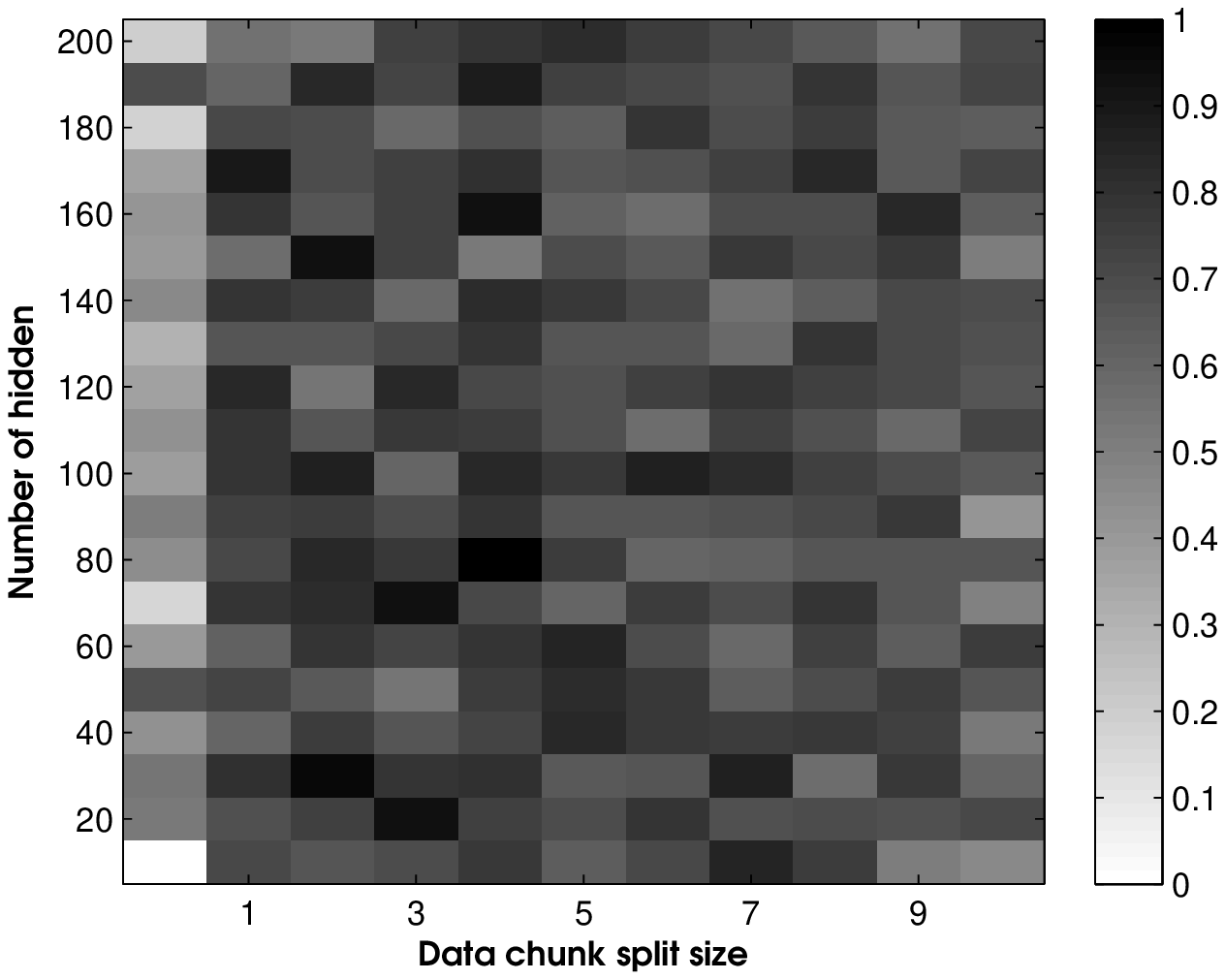}
		\caption{Split size and number of $nh$.}
		\label{fig:pageblocks_map_nh}
	\end{subfigure}
	\begin{subfigure}[b]{0.32\textwidth}
		\includegraphics[width=1\linewidth]{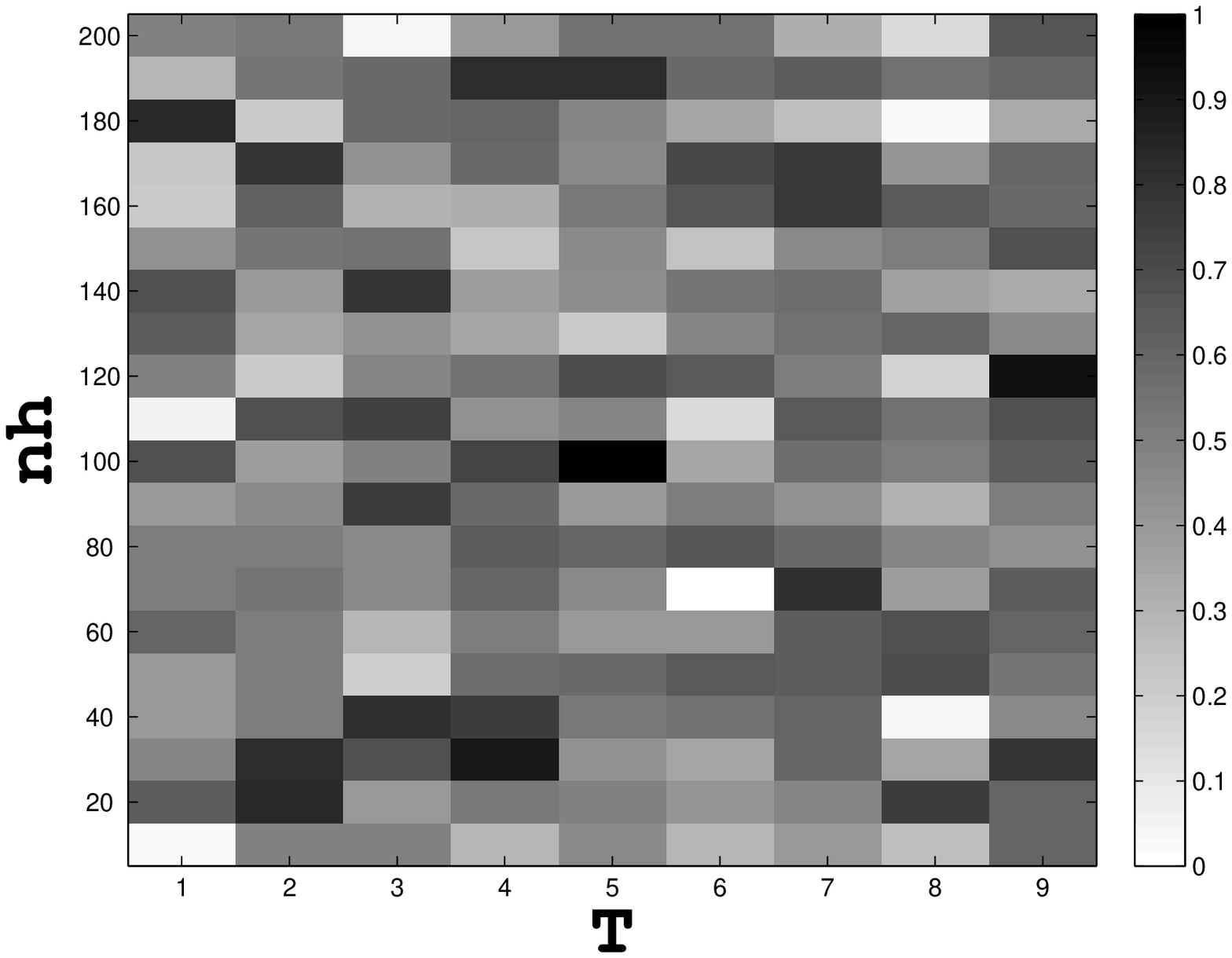}
		\caption{Adaboost $T$ size and number of $nh$.}
		\label{fig:pageblocks_T_nh}
	\end{subfigure}
	\caption{Page blocks data set heatmap.}
	\label{fig:pageblocksres}
\end{figure}

\begin{figure}
	\begin{subfigure}[b]{0.32\textwidth}
		\includegraphics[width=1\linewidth]{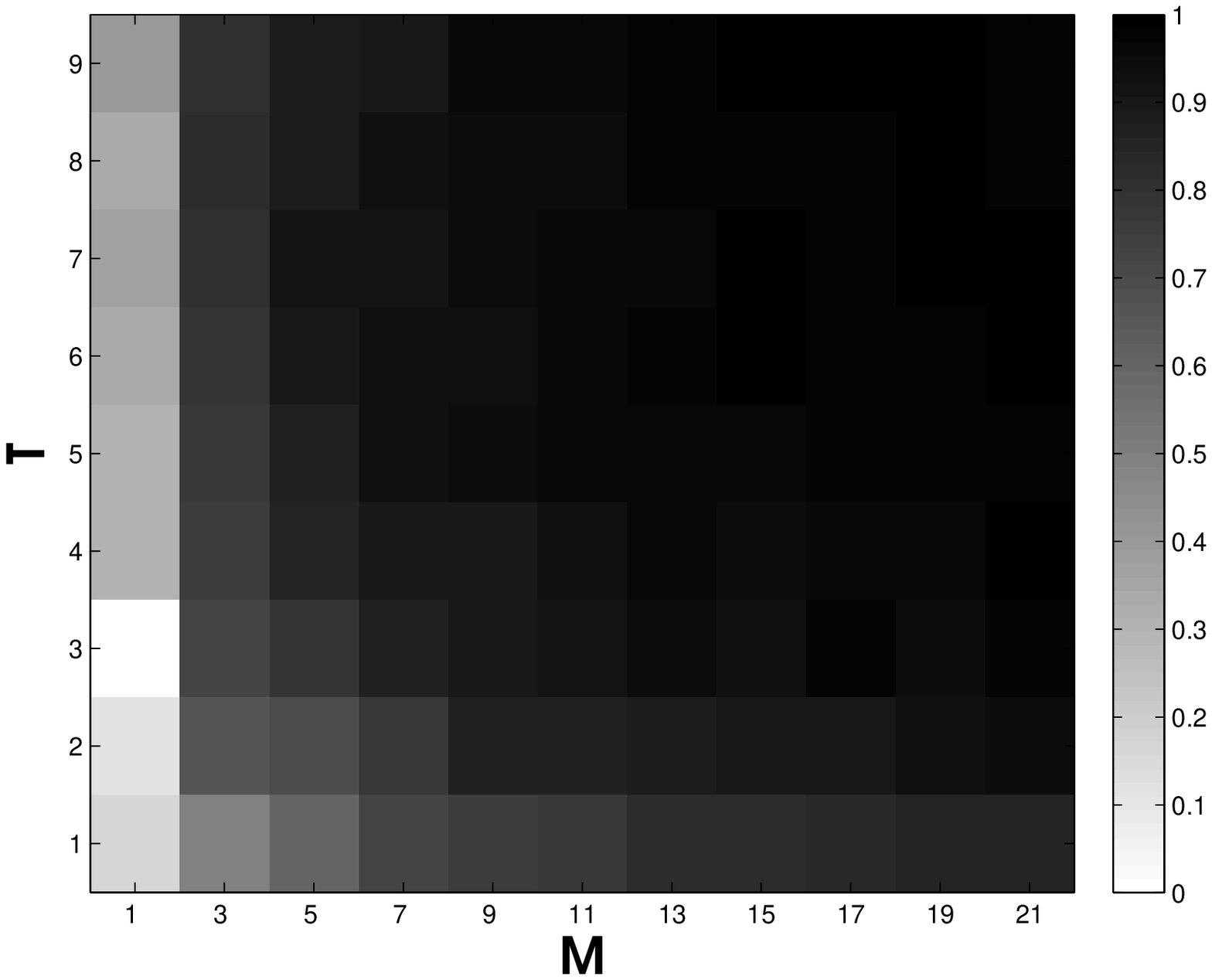}
		\caption{Split size and adaboost $T$ size}
		\label{fig:waveform_map_T}
	\end{subfigure}
	\begin{subfigure}[b]{0.32\textwidth}
		\includegraphics[width=1\linewidth]{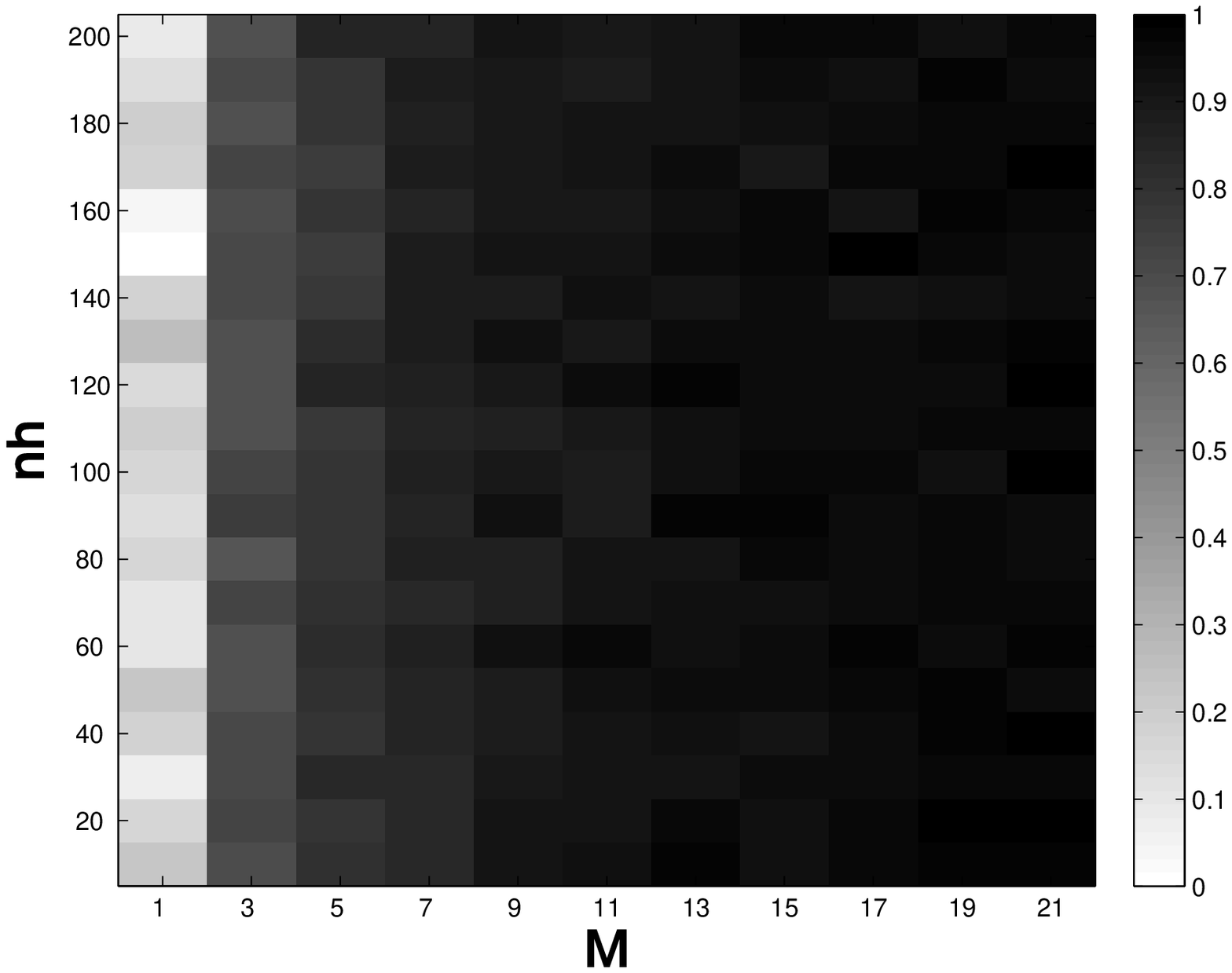}
		\caption{Split size and number of $nh$.}
		\label{fig:waveform_map_nh}
	\end{subfigure}
	\begin{subfigure}[b]{0.32\textwidth}
		\includegraphics[width=1\linewidth]{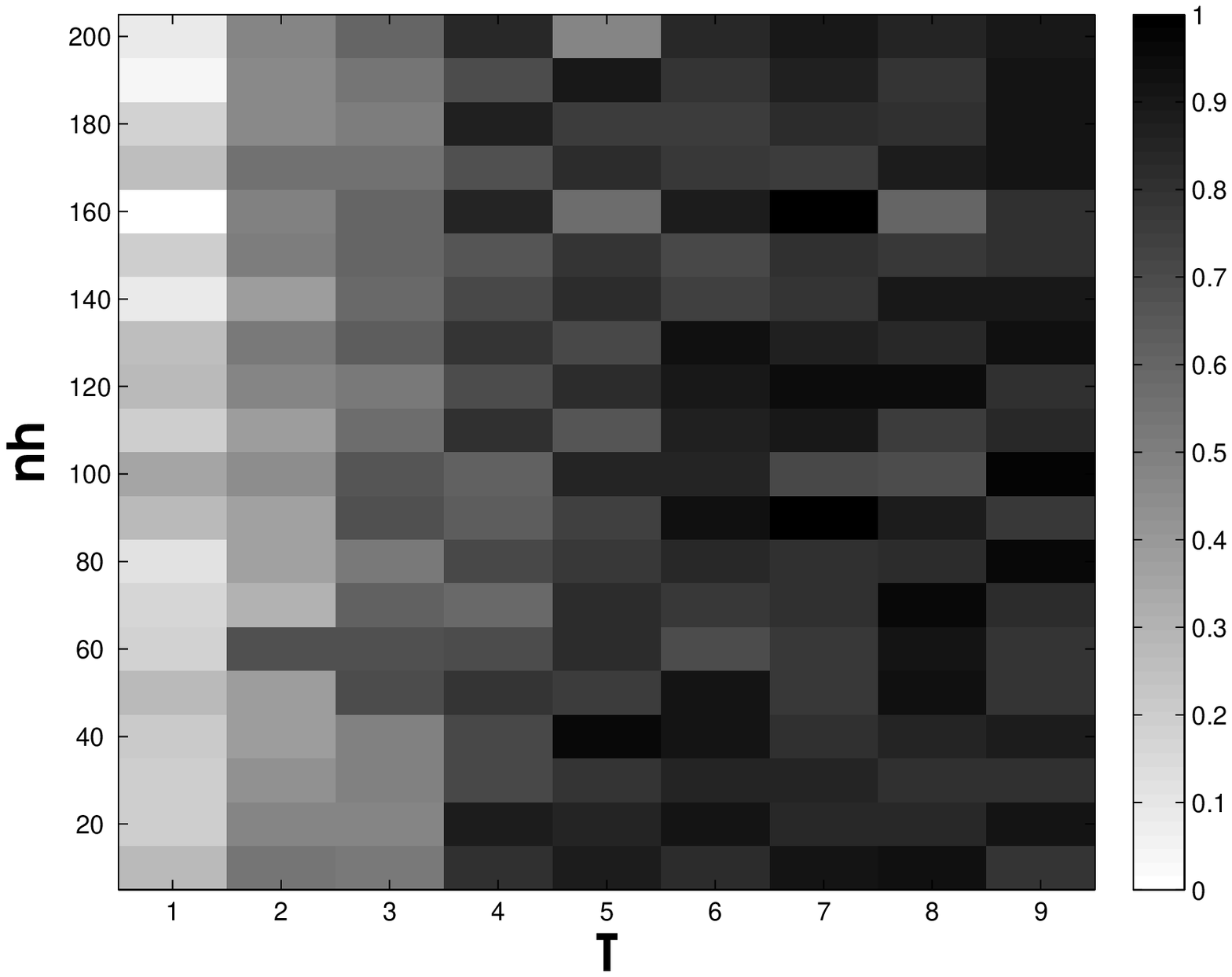}
		\caption{Adaboost $T$ size and number of $nh$.}
		\label{fig:waveform_T_nh}
	\end{subfigure}
	\caption{Waveform data set heatmap.}
	\label{fig:waveformres}
\end{figure}

The results of accuracy and performance tests with real data are shown in Table \ref{tbl:bestres} and Figure \ref{fig:statlogres} - Figure \ref{fig:waveformres}. According to the these results, AdaBoost $T$ size and Mapper size have more impact on the accuracy of ensemble ELM classifier than number of hidden nodes in ELM network.

Accuracy of classification models are visualized by heatmap color coding according to
\begin{itemize}
	\item Map size ($M$) - AdaBoost size ($T$)
	\item Map size ($M$) - Number of hidden nodes ($nh$)
	\item AdaBoost size ($T$) - Number of hidden nodes ($nh$)
\end{itemize}

Figure \ref{fig:statlogres} - Figure \ref{fig:waveformres} are used to plot the quantitative differences in accuracy score of each data set. Heatmaps are two dimensional graphical representations of data with a pre-defined colormap to display values of a matrix \cite{heatmapgenerator}. Heatmaps can be used to understand what parameters affect the accuracy of the classification model. The figures are used to comparatively illustrate accuracy levels across a number of different parameters including Map size, AdaBoost size and the number of hidden nodes in ELM algorithm obtained from the proposed learning method.

\begin{table}
	\caption{Best performance results of data sets}
	\label{tbl:bestres}
	\scriptsize
	\begin{center}
	\begin{tabular}[t]{|c||r|r|r|r|r|r|r|}
		\hline
		Data set & \# C. & T & \# H.N. & Acc & Prec. & Recall & $F_1$ \\
		\hline \hline
		Pendigit & 20 & 10 & 21 & 0,8256 & 0,8369 & 0,8234 & 0,8301 \\
		Skin & 21 & 5 & 21 & 0,9892 & 0,9773 & 0,9913 & 0,9842 \\
		Statlog & 11 & 2 & 21 & 0,9103 & 0,7486 & 0,5069 & 0,6045 \\
		Page Blocks & 1 & 1 & 340 & 0,9404 & 0,9027 & 0,5756 & 0,7030 \\
		Waveform & 19 & 6 & 40 & 0,862 & 0,8680 & 0,8605 & 0,8642 \\
		\hline
	\end{tabular}
	\end{center}
\end{table}

\begin{table}
	\caption{Performance comparison of ELM and proposed model.}
	\label{tbl:comparison}
	\begin{center}
		\begin{tabular}[h]{|c|r|r|r|r|r|}
		\hline
		Data set & Method & $F_1$ & Recall & Precision & Accuracy \\ \hline
		\multirow{2}{*}{Pendigit} & Conventional & 0.8404 & 0.8393 & 0.8416 & 0.8407 \\ 
		& Proposed & 0.8301 & 0.8234 & 0.8369 & 0.8256 \\
		\hline
		\multirow{2}{*}{Skin} & Conventional & 0.9754 & 0.9956 & 0.9583 & 0.9894 \\ 
		& Proposed & 0.9842 & 0.9913 & 0.9773 & 0.9892 \\
		\hline
		\multirow{2}{*}{Statlog} & Conventional & 0.8871 & 0.8556 & 0.9237 & 0.9757 \\
		& Proposed & 0.6045 & 0.5069 & 0.7486 & 0.9103 \\
		\hline
		\multirow{2}{*}{Page-blocks} & Conventional & 0.9873 & 0.9764 & 0.9988 & 0.9977 \\
		& Proposed & 0.7030 & 0.5756 & 0.9027 & 0.9404 \\
		\hline
		\multirow{2}{*}{Waveform} & Conventional & 0.8372 & 0.8368 & 0.8375 & 0.8376 \\
		& Proposed & 0.8642 & 0.8605 & 0.8680 & 0.8620 \\
		\hline
	\end{tabular}
	\end{center}
\end{table}

According to Table \ref{tbl:comparison}, classification performance results of the proposed method have almost the same values with the conventional ELM method.

\subsubsection{Large Scale Classification Data Sets}\label{sec:lsdsacc}
\begin{figure}
		\begin{center}
		\includegraphics[width=0.7\linewidth]{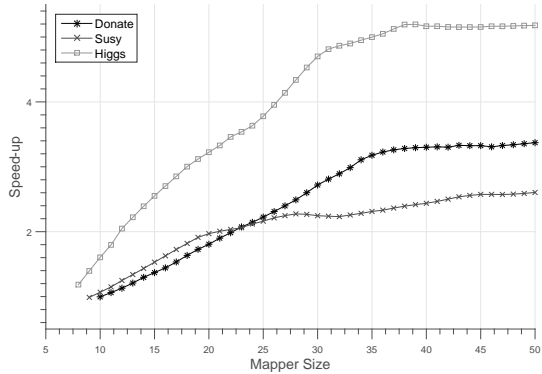}
		\end{center}
		\caption{Stability analysis of ensemble ELM classifiers with Mapper size.}
		\label{fig:ds_large_scale_speed_up}
\end{figure}

Figure \ref{fig:ds_large_scale_speed_up} shows the speed up on mapper size over proposed method on large scale data sets. To asses the effectiveness of the learning algorithm, the time is measured with varying mapper size. Because of high dimensionality, the data sets cannot be trained on a single computer. Then, the standart speed up percentage is modifed such that:
\begin{equation}
	S_p = \frac{t_{\argmin{m} \in M }}{t_p}
\end{equation}
where $t_{\argmin{m} \in M }$ is the total time on minimum mapper that can be achieved to build a classifier model.

As can be seen from the figure, the data sets achives performance improvement in learning time of the algorithm. By examining the trends observed as the number of mappers increases, one can see that \textit{non-linear} speed up is achieved.

\subsection{Stability Analysis}
\begin{figure}
	\begin{subfigure}[b]{0.50\textwidth}
		\includegraphics[width=1\linewidth]{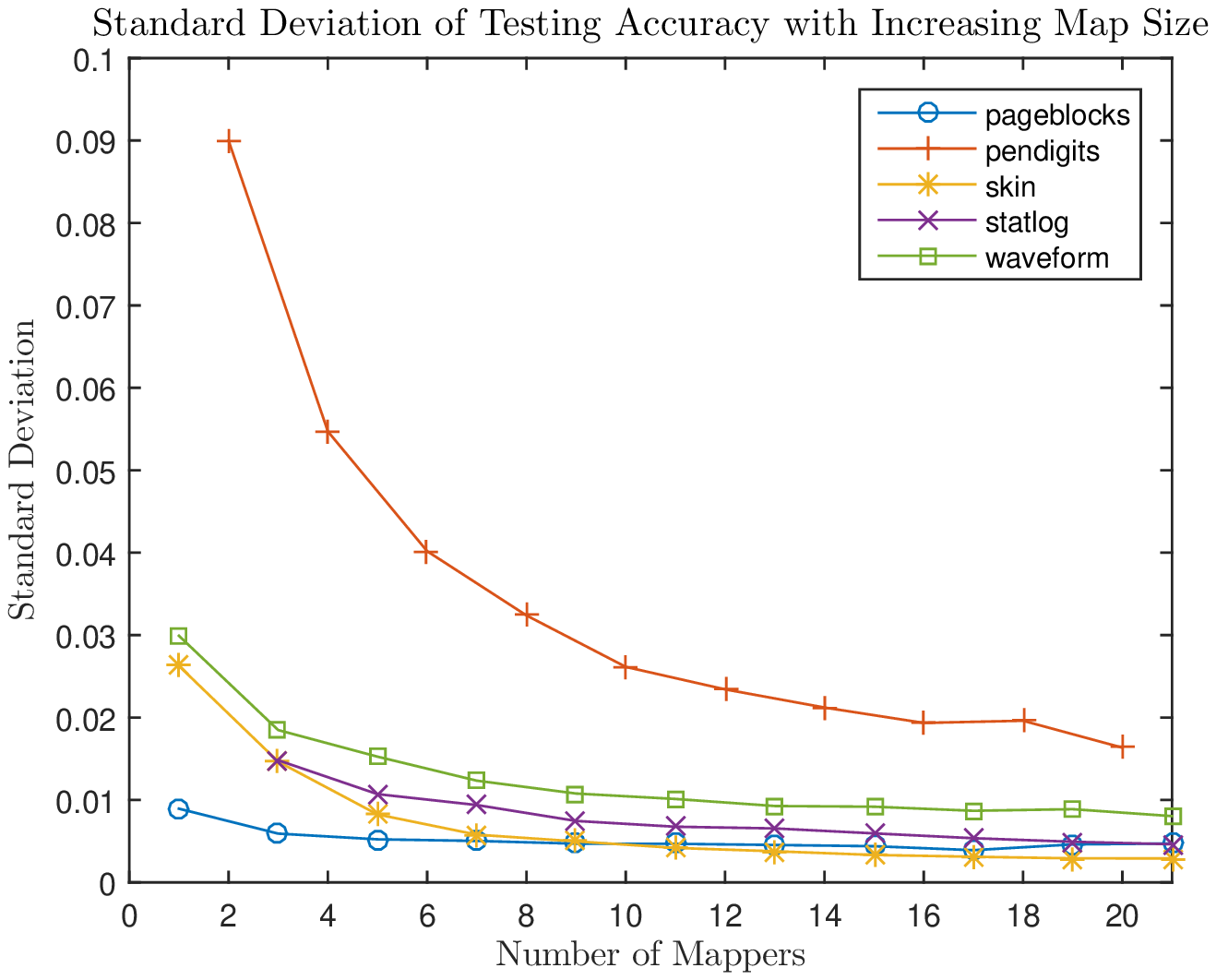}
		\caption{Stability analysis of ensemble ELM classifiers with Mapper size.}
		\label{fig:stability_M}
	\end{subfigure}
	\begin{subfigure}[b]{0.50\textwidth}
		\includegraphics[width=1\linewidth]{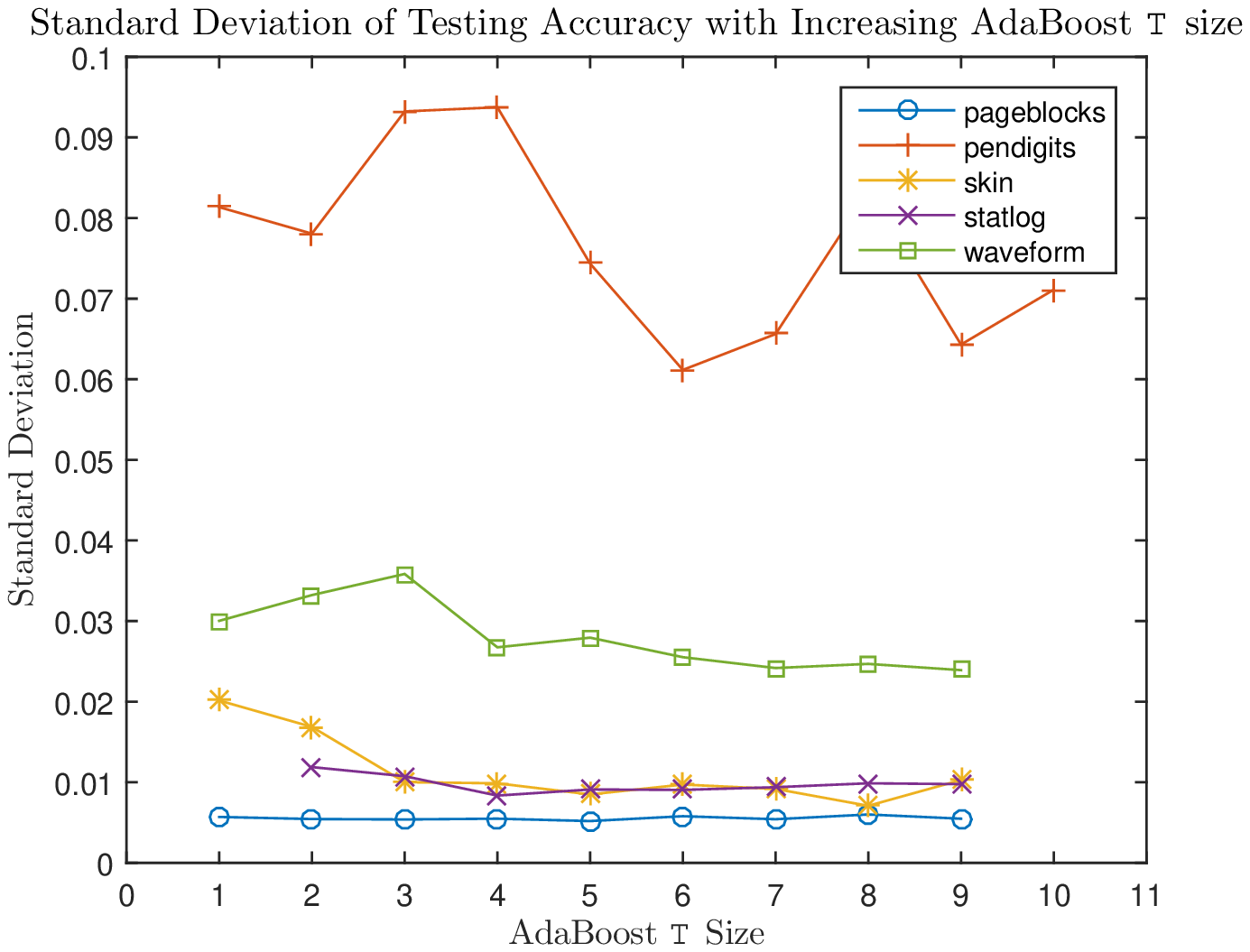}
		\caption{Stability analysis of ensemble ELM classifiers with AdaBoost $T$ size.}
		\label{fig:stability_T}
	\end{subfigure}
	\caption{Stability analysis}
\end{figure}

Standard deviation of testing accuracy of the method is shown in Figure \ref{fig:stability_M} and Figure \ref{fig:stability_T}. We analyzed the stability of ensemble ELM classifier with two aspects, Mapper size and AdaBoost $T$ size. Mapper size is the most important variable for the model stability according to the Figure \ref{fig:stability_M}. From Figure \ref{fig:stability_M} and Figure \ref{fig:stability_T}, we can find that standard deviation of testing accuracy decreases enormously with the increasing of Mapper function size. Through this analysis, one can argue that a model with high Mapper function size do has higher stability than low Mapper function size.
\section{Conclusion and Future Works}\label{sec:conclusion}
In this paper,a parallel AdaBoost extreme learning machine algorithm implementation has been proposed for massive data learning. By creating the overall data set into data chunks, MapReduce based learning algorithm reduces the training time of ELM classification. To overcome the accuracy performance decreasing, distributed ELM is enhanced with AdaBoost method. The experimental results show that AdaBoosted ELM not only reduce the training time of large-scale data sets, but also evaluation metrics of accuracy performance compared with the conventional ELM. 

The proposed AdaBoost based ELM has three different trade-off parameters which are (i) data chunk split size, $M$, (ii) maximum number of iterations, $T$,in AdaBoost Algorithm and lastly (iii) number of hidden layer nodes $nh$ in ELM algorithm. The empirical results in heatmap figures show that parameters $M$ and $T$ are more dominant than parameter $nh$ for the classification accuracy of the hypothesis.

The algorithm is designed to deal with large scale data set ELM training problems. Another objective is to achieve the model's classification performance with same or close to the conventional ELM method. Classification performance results are shown in Section \ref{sec:conv_elm}. The empirical results show us that classification performance results of the proposed method have almost the same values with the conventional ELM method.



%
%

\bibliographystyle{ieeetr}
\bibliography{references}

\begin{thebibliography}{10}

\bibitem{Huang2006489}
G.-B. Huang, Q.-Y. Zhu, and C.-K. Siew, ``Extreme learning machine: Theory and
  applications,'' {\em Neurocomputing}, vol.~70, no.~1–3, pp.~489 -- 501,
  2006.
\newblock Neural Networks Selected Papers from the 7th Brazilian Symposium on
  Neural Networks (SBRN '04) 7th Brazilian Symposium on Neural Networks.

\bibitem{zhao2011xml}
X.-g. Zhao, G.~Wang, X.~Bi, P.~Gong, and Y.~Zhao, ``Xml document classification
  based on elm,'' {\em Neurocomputing}, vol.~74, no.~16, pp.~2444--2451, 2011.

\bibitem{Wang2008262}
G.~Wang, Y.~Zhao, and D.~Wang, ``A protein secondary structure prediction
  framework based on the extreme learning machine,'' {\em Neurocomputing},
  vol.~72, no.~1–3, pp.~262 -- 268, 2008.
\newblock Machine Learning for Signal Processing (MLSP 2006) / Life System
  Modelling, Simulation, and Bio-inspired Computing (LSMS 2007).

\bibitem{Zong20112541}
W.~Zong and G.-B. Huang, ``Face recognition based on extreme learning
  machine,'' {\em Neurocomputing}, vol.~74, no.~16, pp.~2541 -- 2551, 2011.
\newblock Advances in Extreme Learning Machine: Theory and Applications
  Biological Inspired Systems. Computational and Ambient Intelligence Selected
  papers of the 10th International Work-Conference on Artificial Neural
  Networks (IWANN2009).

\bibitem{lan2013extreme}
Y.~Lan, Z.~Hu, Y.~C. Soh, and G.-B. Huang, ``An extreme learning machine
  approach for speaker recognition,'' {\em Neural Computing and Applications},
  vol.~22, no.~3-4, pp.~417--425, 2013.

\bibitem{4470008}
Y.~Lu, V.~Roychowdhury, and L.~Vandenberghe, ``Distributed parallel support
  vector machines in strongly connected networks,'' {\em Neural Networks, IEEE
  Transactions on}, vol.~19, pp.~1167--1178, July 2008.

\bibitem{sun2012study}
Z.~Sun and G.~Fox, ``Study on parallel svm based on mapreduce,'' in {\em
  International Conference on Parallel and Distributed Processing Techniques
  and Applications}, pp.~16--19, Citeseer, 2012.

\bibitem{cloudsvm}
F.~Catak and M.~Balaban, ``Cloudsvm: Training an svm classifier in cloud
  computing systems,'' in {\em Pervasive Computing and the Networked World}
  (Q.~Zu, B.~Hu, and A.~Elçi, eds.), vol.~7719 of {\em Lecture Notes in
  Computer Science}, pp.~57--68, Springer Berlin Heidelberg, 2013.

\bibitem{Panda:2009:PMP:1687553.1687569}
B.~Panda, J.~S. Herbach, S.~Basu, and R.~J. Bayardo, ``Planet: Massively
  parallel learning of tree ensembles with mapreduce,'' {\em Proc. VLDB
  Endow.}, vol.~2, pp.~1426--1437, Aug. 2009.

\bibitem{Zhang:2012:EPK:2247596.2247602}
C.~Zhang, F.~Li, and J.~Jestes, ``Efficient parallel knn joins for large data
  in mapreduce,'' in {\em Proceedings of the 15th International Conference on
  Extending Database Technology}, EDBT '12, (New York, NY, USA), pp.~38--49,
  ACM, 2012.

\bibitem{5372757}
T.~Sun, C.~Shu, F.~Li, H.~Yu, L.~Ma, and Y.~Fang, ``An efficient hierarchical
  clustering method for large datasets with map-reduce,'' in {\em Parallel and
  Distributed Computing, Applications and Technologies, 2009 International
  Conference on}, pp.~494--499, Dec 2009.

\bibitem{6121313}
Y.~He, H.~Tan, W.~Luo, H.~Mao, D.~Ma, S.~Feng, and J.~Fan, ``Mr-dbscan: An
  efficient parallel density-based clustering algorithm using mapreduce,'' in
  {\em Parallel and Distributed Systems (ICPADS), 2011 IEEE 17th International
  Conference on}, pp.~473--480, Dec 2011.

\bibitem{zhao2009}
W.~Zhao, H.~Ma, and Q.~He, ``Parallel k-means clustering based on mapreduce,''
  in {\em Cloud Computing} (M.~Jaatun, G.~Zhao, and C.~Rong, eds.), vol.~5931
  of {\em Lecture Notes in Computer Science}, pp.~674--679, Springer Berlin
  Heidelberg, 2009.

\bibitem{elmstar}
J.~Xin, Z.~Wang, C.~Chen, L.~Ding, G.~Wang, and Y.~Zhao, ``Elm ∗ :
  distributed extreme learning machine with mapreduce,'' {\em World Wide Web},
  vol.~17, no.~5, pp.~1189--1204, 2014.

\bibitem{4012031}
N.-Y. Liang, G.-B. Huang, P.~Saratchandran, and N.~Sundararajan, ``A fast and
  accurate online sequential learning algorithm for feedforward networks,''
  {\em Neural Networks, IEEE Transactions on}, vol.~17, pp.~1411--1423, Nov
  2006.

\bibitem{Sun20112438}
Y.~Sun, Y.~Yuan, and G.~Wang, ``An os-elm based distributed ensemble
  classification framework in \{P2P\} networks,'' {\em Neurocomputing},
  vol.~74, no.~16, pp.~2438 -- 2443, 2011.
\newblock Advances in Extreme Learning Machine: Theory and Applications
  Biological Inspired Systems. Computational and Ambient Intelligence Selected
  papers of the 10th International Work-Conference on Artificial Neural
  Networks (IWANN2009).

\bibitem{Wang2015224}
B.~Wang, S.~Huang, J.~Qiu, Y.~Liu, and G.~Wang, ``Parallel online sequential
  extreme learning machine based on mapreduce,'' {\em Neurocomputing},
  vol.~149, Part A, no.~0, pp.~224 -- 232, 2015.
\newblock Advances in neural networks Selected papers from the Tenth
  International Symposium on Neural Networks (ISNN 2013) Advances in Extreme
  Learning Machines Selected articles from the International Symposium on
  Extreme Learning Machines (ELM 2013).

\bibitem{Bi2015456}
X.~Bi, X.~Zhao, G.~Wang, P.~Zhang, and C.~Wang, ``Distributed extreme learning
  machine with kernels based on mapreduce,'' {\em Neurocomputing}, vol.~149,
  Part A, no.~0, pp.~456 -- 463, 2015.
\newblock Advances in neural networks Selected papers from the Tenth
  International Symposium on Neural Networks (ISNN 2013) Advances in Extreme
  Learning Machines Selected articles from the International Symposium on
  Extreme Learning Machines (ELM 2013).

\bibitem{6565081}
J.~Chen, G.~Zheng, and H.~Chen, ``Elm-mapreduce: Mapreduce accelerated extreme
  learning machine for big spatial data analysis,'' in {\em Control and
  Automation (ICCA), 2013 10th IEEE International Conference on}, pp.~400--405,
  June 2013.

\bibitem{kuncheva2003measures}
L.~I. Kuncheva and C.~J. Whitaker, ``Measures of diversity in classifier
  ensembles and their relationship with the ensemble accuracy,'' {\em Machine
  learning}, vol.~51, no.~2, pp.~181--207, 2003.

\bibitem{Huang06extremelearning}
G.~bin Huang, Q.~yu~Zhu, and C.~kheong Siew, ``Extreme learning machine: A new
  learning scheme of feedforward neural networks,'' in {\em IN PROC. INT. JOINT
  CONF. NEURAL NETW}, pp.~985--990, 2006.

\bibitem{1650244}
G.-B. Huang, L.~Chen, and C.-K. Siew, ``Universal approximation using
  incremental constructive feedforward networks with random hidden nodes,''
  {\em Neural Networks, IEEE Transactions on}, vol.~17, pp.~879--892, July
  2006.

\bibitem{Huang20073056}
G.-B. Huang and L.~Chen, ``Convex incremental extreme learning machine,'' {\em
  Neurocomputing}, vol.~70, no.~16–18, pp.~3056 -- 3062, 2007.
\newblock Neural Network Applications in Electrical Engineering Selected papers
  from the 3rd International Work-Conference on Artificial Neural Networks
  (IWANN 2005) 3rd International Work-Conference on Artificial Neural Networks
  (IWANN 2005).

\bibitem{Huang20083460}
G.-B. Huang and L.~Chen, ``Enhanced random search based incremental extreme
  learning machine,'' {\em Neurocomputing}, vol.~71, no.~16–18, pp.~3460 --
  3468, 2008.
\newblock Advances in Neural Information Processing (ICONIP 2006) / Brazilian
  Symposium on Neural Networks (SBRN 2006).

\bibitem{6866146}
J.~Tang, C.~Deng, G.-B. Huang, and B.~Zhao, ``Compressed-domain ship detection
  on spaceborne optical image using deep neural network and extreme learning
  machine,'' {\em Geoscience and Remote Sensing, IEEE Transactions on},
  vol.~53, pp.~1174--1185, March 2015.

\bibitem{Huang2008576}
G.-B. Huang, M.-B. Li, L.~Chen, and C.-K. Siew, ``Incremental extreme learning
  machine with fully complex hidden nodes,'' {\em Neurocomputing}, vol.~71,
  no.~4–6, pp.~576 -- 583, 2008.
\newblock Neural Networks: Algorithms and Applications 4th International
  Symposium on Neural Networks 50 Years of Artificial Intelligence: a Neuronal
  Approach Campus Multidisciplinary in Perception and Intelligence.

\bibitem{freund1995desicion}
Y.~Freund and R.~E. Schapire, ``A desicion-theoretic generalization of on-line
  learning and an application to boosting,'' in {\em Computational learning
  theory}, pp.~23--37, Springer, 1995.

\bibitem{freund1999short}
Y.~Freund, R.~Schapire, and N.~Abe, ``A short introduction to boosting,'' {\em
  Journal-Japanese Society For Artificial Intelligence}, vol.~14, no.~771-780,
  p.~1612, 1999.

\bibitem{LandesaVazquez2013101}
I.~Landesa-Vázquez and J.~L. Alba-Castro, ``Double-base asymmetric adaboost,''
  {\em Neurocomputing}, vol.~118, no.~0, pp.~101 -- 114, 2013.

\bibitem{Dean:2008:MSD:1327452.1327492}
J.~Dean and S.~Ghemawat, ``Mapreduce: Simplified data processing on large
  clusters,'' {\em Commun. ACM}, vol.~51, pp.~107--113, Jan. 2008.

\bibitem{citeulike:4292765}
M.~C. Schatz, ``{CloudBurst: highly sensitive read mapping with MapReduce.},''
  {\em Bioinformatics (Oxford, England)}, vol.~25, pp.~1363--1369, June 2009.

\bibitem{soft01}
J.~Choi, C.~Choi, B.~Ko, and P.~Kim, ``A method of ddos attack detection using
  http packet pattern and rule engine in cloud computing environment,'' {\em
  Soft Computing}, vol.~18, no.~9, pp.~1697--1703, 2014.

\bibitem{soft02}
M.~Ogiela, A.~Castiglione, and I.~You, ``Soft computing for security services
  in smart and ubiquitous environments,'' {\em Soft Computing}, vol.~18, no.~9,
  pp.~1655--1658, 2014.

\bibitem{bhimji2014hepdoop}
W.~Bhimji, T.~Bristow, and A.~Washbrook, ``Hepdoop: High-energy physics
  analysis using hadoop,'' in {\em Journal of Physics: Conference Series},
  vol.~513, p.~022004, IOP Publishing, 2014.

\bibitem{xu2014privacy}
L.~Xu, H.~Kim, X.~Wang, W.~Shi, and T.~Suh, ``Privacy preserving large scale
  dna read-mapping in mapreduce framework using fpgas,'' in {\em Field
  Programmable Logic and Applications (FPL), 2014 24th International Conference
  on}, pp.~1--4, IEEE, 2014.

\bibitem{661502}
P.~Bartlett, ``The sample complexity of pattern classification with neural
  networks: the size of the weights is more important than the size of the
  network,'' {\em Information Theory, IEEE Transactions on}, vol.~44,
  pp.~525--536, Mar 1998.

\bibitem{Krogh95neuralnetwork}
A.~Krogh and J.~Vedelsby, ``Neural network ensembles, cross validation, and
  active learning,'' in {\em Advances in Neural Information Processing
  Systems}, pp.~231--238, MIT Press, 1995.

\bibitem{libsvmdatasets}
LIBSVM, ``Libsvm data: Classification, regression, and multi-label.''
  http://ntucsu.csie.ntu.edu.tw/

\bibitem{Alimoglu96methodsof}
F.~Alimoglu and E.~Alpaydin, ``Methods of combining multiple classifiers based
  on different representations for pen-based handwritten digit recognition,''
  in {\em Proceedings of the Fifth Turkish Artificial Intelligence and
  Artificial Neural Networks Symposium (TAINN 96}, 1996.

\bibitem{5409447}
R.~Bhatt, G.~Sharma, A.~Dhall, and S.~Chaudhury, ``Efficient skin region
  segmentation using low complexity fuzzy decision tree model,'' in {\em India
  Conference (INDICON), 2009 Annual IEEE}, pp.~1--4, Dec 2009.

\bibitem{Hsu:2002:CMM:2325784.2326870}
C.-W. Hsu and C.-J. Lin, ``A comparison of methods for multiclass support
  vector machines,'' {\em Trans. Neur. Netw.}, vol.~13, pp.~415--425, Mar.
  2002.

\bibitem{Malerba96afurther}
D.~Malerba, F.~Esposito, and G.~Semeraro, ``A further comparison of
  simplification methods for decision-tree induction,'' in {\em In D. Fisher
  and H. Lenz (Eds.), Learning}, pp.~365--374, Springer-Verlag, 1996.

\bibitem{breiman1984classification}
L.~Breiman, J.~Friedman, C.~J. Stone, and R.~A. Olshen, {\em Classification and
  regression trees}.
\newblock CRC press, 1984.

\bibitem{schmidtmann2009evaluation}
I.~Schmidtmann, G.~Hammer, M.~Sariyar, A.~Gerhold-Ay, and
  K.~des~{\"o}ffentlichen Rechts, ``Evaluation des krebsregisters
  nrw--schwerpunkt record linkage,'' {\em Abschlu{\ss}bericht vom}, vol.~11,
  2009.

\bibitem{ucidonation}
UCI, ``Record linkage comparison patterns data set,'' 2011.
\newblock Available online at
  \url{https://archive.ics.uci.edu/ml/datasets/Record+Linkage+Comparison+Patterns}.

\bibitem{baldi2014searching}
P.~Baldi, P.~Sadowski, and D.~Whiteson, ``Searching for exotic particles in
  high-energy physics with deep learning,'' {\em Nature communications},
  vol.~5, 2014.

\bibitem{ucisusy}
UCI, ``Susy data set,'' 2014.
\newblock Available online at
  \url{https://archive.ics.uci.edu/ml/datasets/SUSY}.

\bibitem{ucihiggs}
UCI, ``Higgs data set,'' 2014.
\newblock Available online at
  \url{https://archive.ics.uci.edu/ml/datasets/HIGGS}.

\bibitem{Turpin:2006:UPV:1148170.1148176}
A.~Turpin and F.~Scholer, ``User performance versus precision measures for
  simple search tasks,'' in {\em Proceedings of the 29th Annual International
  ACM SIGIR Conference on Research and Development in Information Retrieval},
  SIGIR '06, (New York, NY, USA), pp.~11--18, ACM, 2006.

\bibitem{Manning:2008:IIR:1394399}
C.~D. Manning, P.~Raghavan, and H.~Sch\"{u}tze, {\em Introduction to
  Information Retrieval}.
\newblock New York, NY, USA: Cambridge University Press, 2008.

\bibitem{Makhoul99performancemeasures}
J.~Makhoul, F.~Kubala, R.~Schwartz, and R.~Weischedel, ``Performance measures
  for information extraction,'' in {\em In Proceedings of DARPA Broadcast News
  Workshop}, pp.~249--252, 1999.

\bibitem{heatmapgenerator}
B.~Khomtchouk, D.~Van~Booven, and C.~Wahlestedt, ``Heatmapgenerator: high
  performance rnaseq and microarray visualization software suite to examine
  differential gene expression levels using an r and c++ hybrid computational
  pipeline,'' {\em Source Code for Biology and Medicine}, vol.~9, no.~1, 2014.

\end{thebibliography}
\end{document}